\documentclass{article}

\PassOptionsToPackage{authoryear,square}{natbib}

\usepackage[main,final]{neurips_2026}

\usepackage[utf8]{inputenc} 
\usepackage[T1]{fontenc}    
\usepackage[colorlinks=true,linkcolor=blue,citecolor=blue,urlcolor=blue]{hyperref}
\usepackage{url}            
\usepackage{booktabs}       
\usepackage{amsfonts}       
\usepackage{nicefrac}       
\usepackage{microtype}      
\usepackage{xcolor}         

\usepackage{amsmath,amssymb}
\usepackage{tabularx}
\usepackage{array}
\newcolumntype{Y}{>{\raggedright\arraybackslash}X}
\usepackage{multirow}
\usepackage{graphicx}
\usepackage{enumitem}
\usepackage{caption}
\usepackage{tikz}
\usepackage{float}
\usepackage{placeins}
\usepackage{algorithm}
\usepackage{algorithmic}
\usetikzlibrary{arrows.meta,positioning,shapes.geometric,fit,calc}
\DeclareRobustCommand{\Letter}{%
  \tikz[baseline=0pt,x=1em,y=1em,line width=0.035em]{%
    \draw (0,0) rectangle (0.9,0.6);
    \draw (0,0) -- (0.9,0.6) (0,0.6) -- (0.9,0);
  }%
}

\setlist[itemize]{leftmargin=1.7em,itemsep=0.18em,topsep=0.22em}
\setlist[enumerate]{leftmargin=1.9em,itemsep=0.18em,topsep=0.22em}
\renewcommand{\arraystretch}{1.12}
\newcommand{\CACC}{\textsc{CACC}}
\newcommand{\CCR}{\textsc{CCR}}
\newcommand{\AMC}{\textsc{AMC}}
\newcommand{\SCH}{\textsc{SCH}}
\newcommand{\SPP}{\textsc{SP}}
\newcommand{\Afirst}{\ensuremath{A_{\mathrm{first}}}}
\newcommand{\Aoracle}{\ensuremath{A_{\mathrm{oracle}}}}
\newcommand{\Avo}{\ensuremath{A_{\mathrm{verifier}\mid\mathrm{oracle}}}}
\newcommand{\Afinal}{\ensuremath{A_{\mathrm{final}}}}
\newcommand{\Choice}{\texttt{choice\_letter}}
\title{Correct but Unselectable: The Hidden Interface Tax in Multi-Candidate Reasoning}

\author{%
  \textbf{Qianli Ma}\textsuperscript{1,2}\quad
  \textbf{Zhiqing Tang}\textsuperscript{2,\Letter}\quad
  \textbf{Fanshuai Meng}\textsuperscript{1,2}\\[0.25em]
  \textbf{Zhi Yao}\textsuperscript{3,2}\quad
  \textbf{Hanshuai Cui}\textsuperscript{3,2}\quad
  \textbf{Weijia Jia}\textsuperscript{2,4,\Letter}\\[0.6em]
  \parbox{0.96\textwidth}{\centering\small
  \textsuperscript{1}Faculty of Arts and Science, Beijing Normal University, Zhuhai 519087, China\\
  \textsuperscript{2}Institute of Artificial Intelligence and Future Networks,\\
  Beijing Normal University, Zhuhai 519087, Guangdong, China\\
  \textsuperscript{3}School of Artificial Intelligence, Beijing Normal University, Beijing 100875, China\\
  \textsuperscript{4}Guangdong Key Lab of AI and Multi-Modal Data Processing,\\
  Beijing Normal-Hong Kong Baptist University, Zhuhai 519087, China\\[0.4em]
  \textsuperscript{\Letter}Correspondence to: Zhiqing Tang (\texttt{zhiqingtang@bnu.edu.cn}),\\
  Weijia Jia (\texttt{jiawj@bnu.edu.cn}).}
}

\date{}

\begin{document}

\maketitle

\begin{abstract}
Small language models deployed under cost, latency, or privacy constraints cannot simply switch to larger models when reasoning fails. Multi-candidate generate-then-rerank offers a natural test-time scaling route. However, more candidates do not always yield higher accuracy. We identify a previously overlooked bottleneck between generation and selection: a correct answer may appear in the candidate pool yet fail to be selected because it is incomplete, unparseable, or misaligned with the answer mode the fixed selector expects. We term this loss the \emph{hidden interface tax} and formalize a \emph{coverage--conversion decomposition} that distinguishes raw oracle coverage from \emph{selectable coverage}, the subset of problems where a correct candidate can actually be consumed by the selector. We propose Compatibility-Aware Candidate Construction (\CACC{}), an interface layer between the candidate source and the fixed selector that repairs fragments, aligns answer modes, and removes selector-confusing noise without retraining either component. On numeric reasoning (GSM8K, \texttt{competition\_math}), \CACC{} raises both oracle coverage and final accuracy. On GPQA Diamond, \CACC{} alone improves final accuracy from 5.6\% to 19.2\% (+13.6\,pp); combined with proposer-side strengthening it reaches 25.8\% (+20.2\,pp). On MMLU-Pro, \CACC{} exposes a coverage--conversion gap: answer-mode-compatible construction raises oracle coverage, but final accuracy decreases by 0.85\,pp under the original selector; adding proposer-side strengthening improves final accuracy.

\end{abstract}

\section{Introduction}

Small-model reasoning systems are deployed in local assistants, intranet question answering, code/math copilots, and edge offline applications. In these scenarios, model invocation cost, response latency, data privacy, and deployment controllability constrain system design more directly than a single benchmark score. Compared to large models, small models are better suited to such constrained deployments, but they are also more likely to produce incomplete reasoning, wrong final answers, format drift, and inconsistent answer modes on multiple-choice questions (MCQ). Multi-candidate generate-then-rerank has emerged as a natural test-time scaling route for these settings, where additional sampling under a fixed generator and selector is expected to raise accuracy.

\begin{figure}[t]
\centering
\includegraphics[width=\linewidth]{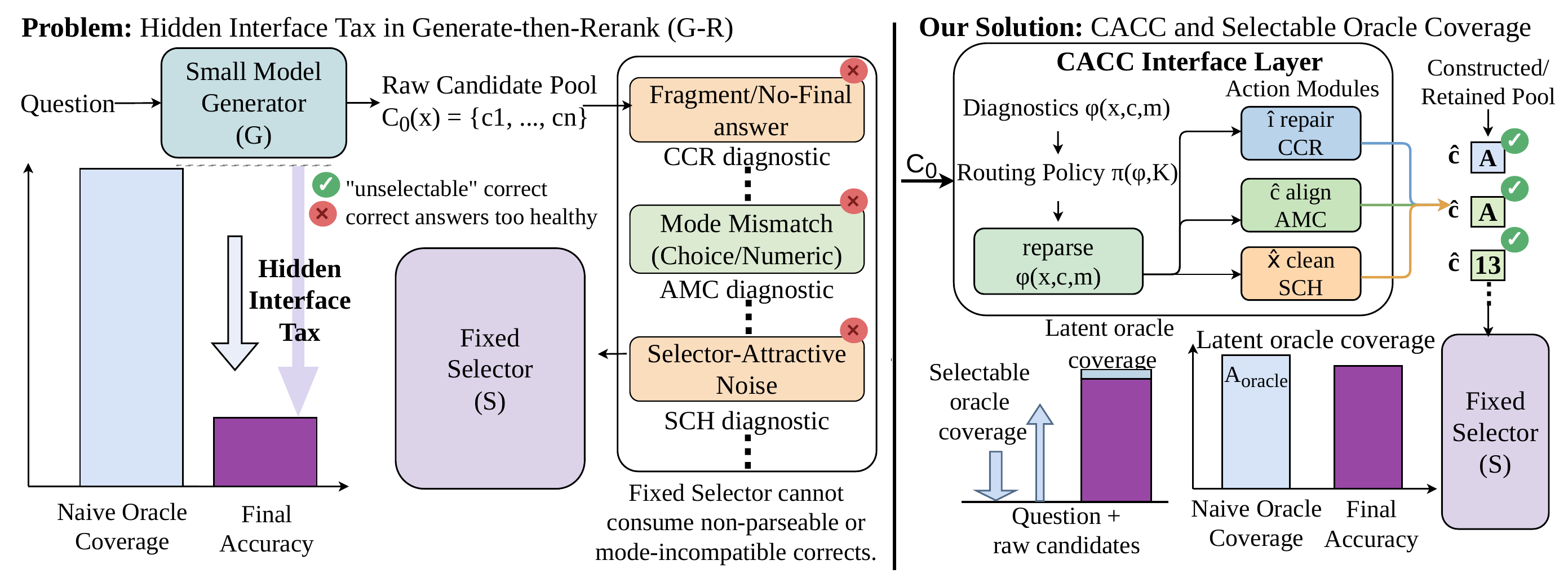}
\caption{The missing candidate-interface layer in small-model multi-candidate reasoning. Existing methods optimize the candidate source or the selector, while whether the pool contains parseable, answer-mode-compatible, and selector-consumable candidates remains an independent bottleneck.}
\label{fig:challenge}
\end{figure}

However, increasing the number of candidates does not automatically yield higher accuracy. Added candidates may be fragments, partial solutions, unparseable strings, or wrong answers that nonetheless attract the verifier. On MCQ tasks, candidates must produce the benchmark-mandated choice letter. Otherwise, correct underlying reasoning still fails at readout. These failure modes arise not at the candidate source or the selector, but at the interface between them (Figure~\ref{fig:challenge}). Existing test-time scaling and self-consistency methods focus on expanding the candidate source \citep{wang2023selfconsistency,snell2025scaling,brown2024monkeys}, while verifiers and rerankers focus on improving selection \citep{cobbe2021training,lightman2024verify}, but neither addresses the candidate-interface layer that determines whether a correct candidate can actually be consumed by the fixed selector. The resulting loss from raw coverage to final accuracy constitutes a \emph{hidden interface tax} that existing methods neither measure nor reduce. To address this, two challenges must be resolved.

\textit{The first challenge is how to identify and quantify the interface bottleneck separately from source quality and selector quality.} Standard practice uses oracle coverage as the upper bound for final accuracy, measuring only whether any candidate in the pool is correct. Final accuracy alone conflates two distinct failure modes: the correct candidate is absent from the pool (a coverage failure), and the correct candidate is present but cannot be consumed by the fixed selector (a conversion failure). Without separating these, practitioners cannot diagnose whether to add more candidates or to fix interface alignment. Existing test-time scaling analyses track candidate count, sampling strategy, or reward-model quality \citep{wang2023selfconsistency,snell2025scaling,brown2024monkeys,liu2025computeoptimal,sinha2026diminishingreturns}, but do not isolate the interface as an independent bottleneck.

\textit{The second challenge is how to construct interface-compatible candidates under a fixed selector without accessing gold labels at test time.} Interface failures span multiple types: incomplete reasoning before the final answer (fragments), answer-mode mismatch where a numeric answer appears in an MCQ slot or a choice letter is missing, and selector-confusing noise such as scaffold residue and near-duplicate wrappers. Trajectory-improvement methods such as Self-Refine and Reflexion alter the generation process itself \citep{madaan2023selfrefine,shinn2023reflexion}, and verifiers or process supervisors choose among candidates \citep{cobbe2021training,lightman2024verify,luo2024automatedprocess}, but neither repairs the interface state of already-generated candidates. The construction mechanism must therefore operate between the fixed generator and the fixed selector, using only gold-free signals at inference time while drawing on offline-calibrated rules for routing and repair.

In this paper, we observe that the candidate interface between generation and selection is an independent, optimizable layer. We propose Compatibility-Aware Candidate Construction (\CACC{}), which targets the distribution of candidates that reaches the fixed selector rather than improving the generator or the selector themselves. This layer-level separation explains why methods that raise oracle coverage sometimes fail to raise final accuracy, and why interface-side repair can succeed where further sampling cannot. On GPQA Diamond, \CACC{} alone improves final accuracy from 5.6\% to 19.2\% (+13.6\,pp).

The first two contributions address the measurement and construction challenges respectively. The main contributions are as follows.
\begin{enumerate}
  \item \textbf{Coverage--conversion decomposition and selectable coverage.} We formalize the coverage--conversion gap in generate-then-rerank inference, defining \emph{selectable coverage} to distinguish ``the correct candidate has appeared'' from ``the correct candidate can be consumed and selected by the fixed selector.'' This decomposition separates coverage and conversion failures, and relates selectable coverage to final accuracy under a fixed selector.
  \item \textbf{Compatibility-Aware Candidate Construction (\CACC{}).} We introduce \CACC{}, an interface layer combining offline rule calibration with online gold-free candidate construction. The offline stage freezes a diagnostic rulebook from benchmark schemas, parser contracts, and audited failure cases. The online stage executes \CCR{} (fragment repair), \AMC{} (answer-mode alignment), and \SCH{} (candidate hygiene) under the frozen policy, keeping the existing selector unchanged.
  \item \textbf{Empirical effectiveness.} Under matched-budget settings, \CACC{} raises both oracle and final accuracy on numeric reasoning (GSM8K, \texttt{competition\_math}). On MCQ, composing \CACC{} with proposer-side strengthening (\SPP{}) reaches $+20.2$\,pp on GPQA Diamond and narrows the coverage--conversion gap on MMLU-Pro. Second-family paired closure on both benchmarks supports recoverable compatibility across generator families.
\end{enumerate}

\section{Related Work}

\textbf{Candidate-source expansion: multi-sampling, self-consistency, and search.}
Self-consistency, multi-sampling, Tree-of-Thought (ToT), and search-based reasoning enlarge candidate sources to raise the probability that a correct trajectory appears \citep{wei2022cot,kojima2022zeroshot,wang2023selfconsistency,yao2023tot,yao2023react,gao2023pal,lewkowycz2022minerva,zelikman2024quietstar,zhou2024selfdiscover,zhou2024lats,besta2024got,koh2024treesearch,xie2024mcts,zhao2024marco}. Recent test-time scaling work analyzes compute allocation, reward model, and problem difficulty \citep{brown2024monkeys,snell2025scaling,liu2025computeoptimal,muennighoff2025s1,sinha2026diminishingreturns}. These methods expand source-side coverage but do not constrain whether candidates are complete, parseable, or answer-mode compatible, nor guarantee that a fixed verifier can select the correct candidate from the enlarged pool. Power Sampling shows that distribution shaping can raise raw oracle coverage yet hurt final accuracy when misaligned with the verifier, confirming that these are upstream extensions rather than same-layer alternatives to \CACC{}.

\textbf{Trajectory refinement and test-time feedback.}
Self-Refine, Reflexion, self-rewarding, code-aided reasoning, fine-tuning at test time (FTTT), and reasoning-model post-training improve trajectories via feedback, self-correction, or test-time optimization \citep{madaan2023selfrefine,shinn2023reflexion,yuan2024selfrewarding,li2024chaincode,gou2024tora,kumar2024selfcorrect,wang2024selfcorrection,li2025fttt,kimiteam2025k15,ye2025limo,openthoughts2025data,song2025rewardenough}. These methods alter the generation process or model state rather than the interface between existing candidates and the selector. \CACC{} occupies a different position in the system stack, consuming outputs from the candidate source and performing fragment repair, answer-mode alignment, and candidate hygiene before the fixed selector.

\textbf{Verifier/reranker and the candidate interface.}
Verifiers, rerankers, and preference optimization select the final answer from a candidate pool \citep{christiano2017preferences,ouyang2022training,bai2022constitutional,rafailov2023dpo,cobbe2021training,uesato2022solving,lightman2024verify,zheng2023judging,lambert2024rewardbench,liu2024rmbench}. Recent work extends step-level checking, process supervision, thought reuse, and automated verifiers to math and general reasoning \citep{miao2024selfcheck,wang2024mathshepherd,setlur2024rewarding,luo2024automatedprocess,zheng2024processbench,ahmed2026retrievalthought}. However, a selector cannot rescue an oracle miss, and even when a correct candidate is present, candidate format, duplication, answer mode, and final-answer wrappers still alter selection behavior. The verifier's selection rate among oracle-hit candidates is therefore a joint outcome of the selector and the candidate interface, and \CACC{} improves this input distribution without replacing the verifier.

Our experiments evaluate on numeric open-ended (GSM8K, \texttt{competition\_math:numeric}) and MCQ (MMLU-Pro, GPQA Diamond) benchmarks \citep{cobbe2021training,hendrycks2021math,hendrycks2021mmlu,rein2023gpqa,wang2024mmlupro}, contextualized against a broader set of reasoning benchmarks \citep{chen2024livebench,zhang2024gsm1k,mirzadeh2024gsmsymbolic,he2024olympiadbench,gao2024omnimath,glazer2024frontiermath} and the open small-model ecosystem \citep{qwen2024qwen25,yang2024qwen25math,qwen2025qwen3,dubey2024llama3,team2024gemma,jiang2023mistral,abdin2024phi3,shao2024deepseekmath,deepseek2024v3,guo2025deepseekr1,kimiteam2025k15,qwen2024qwq,openai2024o1}. None of these benchmarks or models includes an explicit candidate-interface evaluation, which is the gap \CACC{} addresses.

\section{Problem Formulation}
\label{sec:problem}

This section formalizes the candidate interface in generate-then-rerank inference. We modify neither the small-model parameters nor the final selector. The focus is the interface state between the generator's outputs and the selector's inputs. The candidate source generates several candidates from the question, context, and possible options. The candidate-construction layer then repairs, aligns the answer mode, and cleans the raw candidates. Finally, the fixed selector picks one candidate from the retained pool, and a benchmark parser or scoring endpoint determines the final answer. Our central question is: under a fixed selector, can the candidate interface convert ``the correct candidate has appeared'' into ``the correct candidate can be selected''?

For a sample in dataset $\mathcal D$, let $x$ denote the input instance, $y_x$ the ground-truth answer, and $m$ the benchmark-locked answer mode (e.g.\ numeric or \Choice{}). The generator first produces a raw candidate pool:
\begin{equation}
\label{eq:raw-pool}
C_0(x)=\{c_1,\ldots,c_n\}.
\end{equation}
The candidate-construction layer maps $C_0$ to a retained pool
\begin{equation}
\label{eq:cacc-pool}
C_K(x,m)=R_K\!\left(T_m(C_0(x),x)\right),
\end{equation}
where $T_m$ is a mode-aware candidate-construction operator and $R_K$ controls deduplication, ranking, and the retained cap. The fixed selector $S$ chooses
\begin{equation}
\label{eq:selector}
\hat{c}=S(x,C_K), \qquad \hat{y}=\mathrm{parse}_m(\hat{c}).
\end{equation}
Here, $\mathrm{parse}_m(\cdot)$ denotes the benchmark-specific answer parser under answer mode $m$, e.g., numeric extraction for open-ended tasks and choice-letter extraction for MCQ tasks.

Our key constraint is the fixed selector. Within each Base/CACC comparison we modify only $T_m$ or the candidate source, keeping the final selection function $S$ unchanged.

We report four core quantities:
\begin{align}
\Afirst &=
\frac{1}{|\mathcal D|}
\sum_{x\in\mathcal D}
\mathbb{I}\!\left\{\mathrm{parse}_m(c_1)=y_x\right\},\\
\Aoracle &=
\frac{1}{|\mathcal D|}
\sum_{x\in\mathcal D}
\mathbb{I}\!\left\{\exists c\in C_K(x,m):\mathrm{parse}_m(c)=y_x\right\},\\
\Afinal &=
\frac{1}{|\mathcal D|}
\sum_{x\in\mathcal D}
\mathbb{I}\!\left\{\mathrm{parse}_m(S(x,C_K))=y_x\right\},\\
\Avo &=
\Pr\!\left[
\mathrm{parse}_m(S(x,C_K))=y_x
\,\middle|\,
\exists c\in C_K(x,m):\mathrm{parse}_m(c)=y_x
\right].
\end{align}
where $\mathbb{I}\{\cdot\}$ denotes the indicator function, which equals 1 when the condition holds and 0 otherwise. We abbreviate $\Avo$ as V|O in tables and figures throughout.

For an extractive selector, $\Afinal = \Aoracle \times \Avo$ exactly, since a correct final answer requires a correct candidate in the pool. This decomposition separates two distinct bottlenecks: in a coverage-limited regime the pool lacks a correct answer, whereas in a compatibility-limited regime the correct answer is present but cannot be selected by the fixed selector.

\textbf{Selectable coverage.}
Plain oracle coverage only requires that a parseable correct answer appear in the pool. \emph{Selectable coverage} additionally requires that the correct candidate lie within a distribution the fixed selector can consume. Formally, define the \emph{selectable-correct hit} event
\begin{equation}
\label{eq:hsel}
\begin{aligned}
H_{\mathrm{sel}}(C_K,x,m)
= \mathbb{I}\!\left\{
\begin{array}{l}
\exists c\in C_K(x,m):\ 
\mathrm{Correct}(c,y_x)\wedge \mathrm{Parseable}(c,m)\\
\qquad\qquad\wedge\ \mathrm{ModeMatch}(c,m)\wedge \mathrm{Clean}(c)
\end{array}
\right\}.
\end{aligned}
\end{equation}
where $\mathrm{Correct}$ requires semantic correctness, $\mathrm{Parseable}$ requires the benchmark parser to read the candidate, $\mathrm{ModeMatch}$ requires the candidate to follow the answer mode $m$ (a numeric span or \Choice{}), and $\mathrm{Clean}$ requires the absence of selector-attractive noise such as instruction residue, scaffold leakage, or Invalid finals. This yields the \emph{selectable oracle coverage}
\begin{equation}
\label{eq:aselo}
\Aoracle^{\mathrm{sel}} = \Pr\!\left[H_{\mathrm{sel}}(C_K,x,m)=1\right].
\end{equation}
Let $h=\Aoracle^{\mathrm{sel}}$ and $q_b=\Pr[\mathrm{parse}_m(S(x,C_K))=y_x\mid H_{\mathrm{sel}}=b]$. The law of total probability gives the exact identity
\begin{equation}
\label{eq:aselfinal}
\Afinal=hq_1+(1-h)q_0.
\end{equation}
For a zero-probability branch, its weighted contribution is defined as zero. Writing $r=(1-h)q_0$ and $\Avo^{\mathrm{sel}}=q_1$, we obtain
\begin{equation}
\label{eq:residual-gap}
\Aoracle-\Afinal=(\Aoracle-h)+h(1-q_1)-r.
\end{equation}
When $r=0$, $h$ is guaranteed to be an upper bound; with a nonzero residual it need not be: an imperfect diagnostic predicate can reject a candidate that the selector nevertheless answers correctly. For an extractive selector with a common parser, raw oracle coverage remains an upper bound. The compatibility predicate is an operational diagnostic, while the residual accounts for its incomplete agreement with selection. We report oracle coverage, conditional conversion, and final accuracy directly; paired interventions in Section~\ref{sec:controls} measure specific interface effects.

\paragraph{Worked example.}
On MMLU-Pro example \texttt{mmlu-pro-test-00072} (gold J), the recorded eight-candidate \CACC{} pool parses to $[\text{--},A,\text{--},A,\text{--},A,J,F]$. Candidate 7 contains \texttt{Final answer: J}, so oracle coverage is satisfied; majority readout selects A (recorded slot 2), so final correctness is zero. This post-hoc pool example makes the coverage--conversion distinction directly checkable.

\section{Method}
\label{sec:method}

\begin{figure}[t]
\centering
\includegraphics[width=\linewidth]{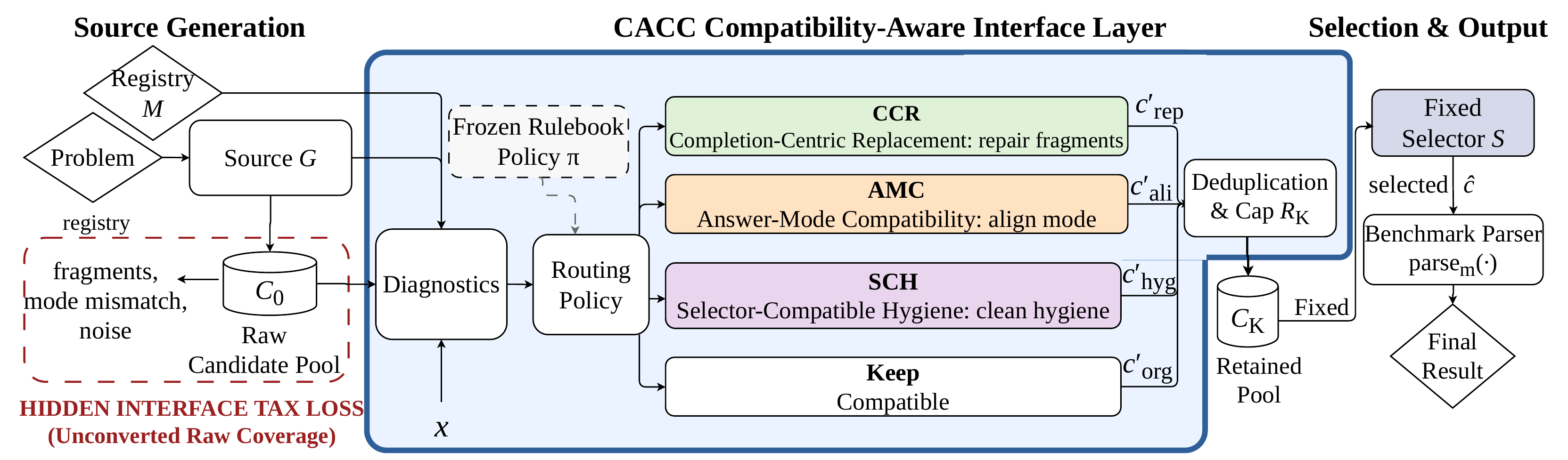}
\caption{The offline--online flow of \CACC{}. Lower half: offline calibration produces a frozen rulebook and routing policy. Upper half: online inference applies gold-free diagnostics and construction to the raw candidate pool before the fixed selector.}
\label{fig:method}
\end{figure}

\CACC{} is a two-stage candidate-interface system (Figure~\ref{fig:method}). The offline stage freezes a diagnostic rulebook from benchmark schemas and audited failure cases. The online stage applies the frozen rules to repair, align, and clean the raw candidate pool before it reaches the fixed selector. The coverage--conversion decomposition (Section~\ref{sec:problem}) identifies where the interface bottleneck lies, and the construction pipeline described here targets it. The basic composition is
\begin{equation}
\label{eq:cacc-modules}
\CACC = \CCR + \AMC + \SCH .
\end{equation}

\textbf{Offline calibration.}
The offline module (lower half of Figure~\ref{fig:method}) takes benchmark schemas, parser configurations, prompt/candidate traces, scoring manifests, and a small set of audited failure cases as input, and outputs a frozen rulebook $\mathcal R$ and routing policy $\pi$. Failure attribution and threshold calibration may use dev or audit-split labels in this stage, but online inference reads no gold labels and never modifies rules according to test-set labels. Concretely, the offline stage first builds the answer-mode registry $\mathcal M$ and parser contract. It then defines candidate diagnostic features $\phi(x,c,m)$ covering parseability, completion, Answer-mode mismatch, wrapper consistency, residue, scaffold leakage, and selector-attractive noise. Next, it instantiates action rule templates for \CCR{}/\AMC{}/\SCH{}. Finally, it calibrates the routing policy $\pi:\phi(x,c,m)\mapsto a\in\{\text{keep, repair, reparse, drop, replace}\}$ on the audit split. The MMLU-Pro \texttt{CCR-only}/no-\AMC{} provenance in Table~\ref{tab:table4} shows that, without locking the answer-mode registry and parser contract, MCQ candidates are severely underestimated under an incorrect numeric readout; \AMC{} is therefore a benchmark contract written into the rulebook offline, rather than an after-the-fact formatting step.

\textbf{Online inference.}
The online module (upper half of Figure~\ref{fig:method}) samples a raw candidate pool $C_0$ under a fixed budget, computes gold-free observable signals $\phi$ (parser status, answer mode, final-answer marker, residue/wrapper, conflict, verifier score/margin) for each candidate, and routes them by the frozen policy $\pi$ to buckets such as incomplete/no-final, answer-mode/parser mismatch, and residue/conflict/verifier-distractor; \CCR{}/\AMC{}/\SCH{} actions are then applied per bucket. The system forms a union of original and constructed candidates, then applies deduplication, retained-cap selection, and a parser check to obtain $C_K$, which is finally passed to the unchanged fixed selector $S$. The three modules differ in where they intervene: \CCR{} primarily affects coverage (when correct answers are missing), \AMC{} primarily affects conversion (when candidates are incompatible with the answer mode), and \SCH{} primarily improves selector input distribution (when correct candidates are drowned by noise and duplication). Algorithm~\ref{alg:cacc} states the full policy-parameterized procedure. The offline rulebook components used in our implementation are summarized in Appendix Table~\ref{tab:app-rulebook}.

\begin{algorithm}[t]
\caption{Offline--online \CACC{}}
\label{alg:cacc}
\renewcommand{\algorithmicrequire}{\textbf{Input:}}
\renewcommand{\algorithmicensure}{\textbf{Output:}}
\begin{algorithmic}[1]
\REQUIRE Benchmark schemas, parser configs, audited traces (offline); problem $x$, raw pool $C_0$, retained cap $K$, fixed selector $S$ (online)
\ENSURE Selected answer $\hat y$, retained pool $C_K$
\STATE \textbf{// Offline calibration}
\STATE Build answer-mode registry $\mathcal M$ and diagnostics $\phi$
\STATE Instantiate rulebook $\mathcal R=\{\CCR\text{-rules},\AMC\text{-rules},\SCH\text{-rules}\}$
\STATE Calibrate and freeze routing policy $\pi(\phi,K)$ on audited traces
\STATE \textbf{// Online inference}
\STATE Infer $m\leftarrow\mathcal M(x)$; diagnose $D=\{\phi(x,c,m):c\in C_0\}$
\STATE Apply frozen rules: $C' \leftarrow \mathrm{apply}(C_0,D,\pi,\mathcal R)$
\STATE Retain: $C_K \leftarrow \mathrm{retain}_K(\mathrm{dedup}(\mathrm{parsecheck}(C',m)),K)$
\STATE Select $\hat c \leftarrow S(x,C_K)$ \COMMENT{Selector unchanged}
\STATE Parse $\hat y \leftarrow \mathrm{parse}_m(\hat c)$
\RETURN $(\hat y, C_K)$
\end{algorithmic}
\end{algorithm}

\textbf{Implementation details.}
In our reference implementation the routing policy $\pi$ is a deterministic decision list parameterized by the diagnostic vector $\phi$ and a fixed retained-cap budget $K$. \CCR{} fires when at least one in-pool candidate carries a fragment or missing-final span and at least one slot falls below the parser's completion threshold, using completion-oriented generation before merging and fixed-cap retention. \AMC{} fires whenever the answer-mode registry $\mathcal M(x)$ disagrees with a candidate's final-answer wrapper; for MCQ tasks it enforces \Choice{} normalization on top of any non-letter answer, while for numeric tasks it normalizes the final numeric span and protects boxed-answer wrappers. \SCH{} fires when residue, scaffold leakage, or Wrapper drift triggers the verifier's distractor signal; deduplication is applied last so that the retained cap is filled with mode-aligned, maximally diverse candidates rather than near-duplicates of a dominant scaffold. \CCR{}, \AMC{}, and \SCH{} use deterministic rules (regex matching, span extraction, template filling); completion additionally calls the generator and requires rescoring. \CACC{} keeps the retained cap fixed without retraining the verifier, but total inference cost is not matched. The full rulebook components and parser-binding contracts are reported in Appendix~\ref{app:algorithm} and Appendix Table~\ref{tab:app-parser-rules}; completion settings and costs are in Appendix~\ref{app:controls}.

\section{Experiments and Mechanistic Analysis}

\textbf{Setup.}
All main experiments follow a generate-then-rerank protocol. Each comparison holds its retained cap, parser, scoring endpoint, and downstream selector fixed. The evidence spans three categories. Numeric open-ended reasoning (GSM8K / \texttt{competition\_math:numeric}) evaluates direct final gains. Interface stress tests (MMLU-Pro / GPQA Diamond) probe MCQ answer-mode alignment, candidate hygiene, and bounded generator-family recovery. Composition evidence asks whether the coverage--conversion gap can be systematically changed by proposer-side strengthening, conversion-aware rescue, and coverage-pressure interventions.

\textbf{Models and inference stack.}
The main pipeline uses Qwen3-1.7B to generate the raw candidate pool, \CACC{} to process the candidate interface, and a trained verifier-reranker held fixed across comparisons to select the final answer. \CACC{}-only and matched baselines share the same generator, retained cap, parser, and verifier. The \CACC{}+\SPP{} rows use a Qwen3-8B proposer / prompt-mixture as proposer-side strengthening and are reported only as selected-regime composition evidence rather than folded into the canonical \CACC{} result. Full checkpoint ids are listed in Appendix~\ref{app:protocol}.

\begin{figure}[t]
\centering
\includegraphics[width=\linewidth]{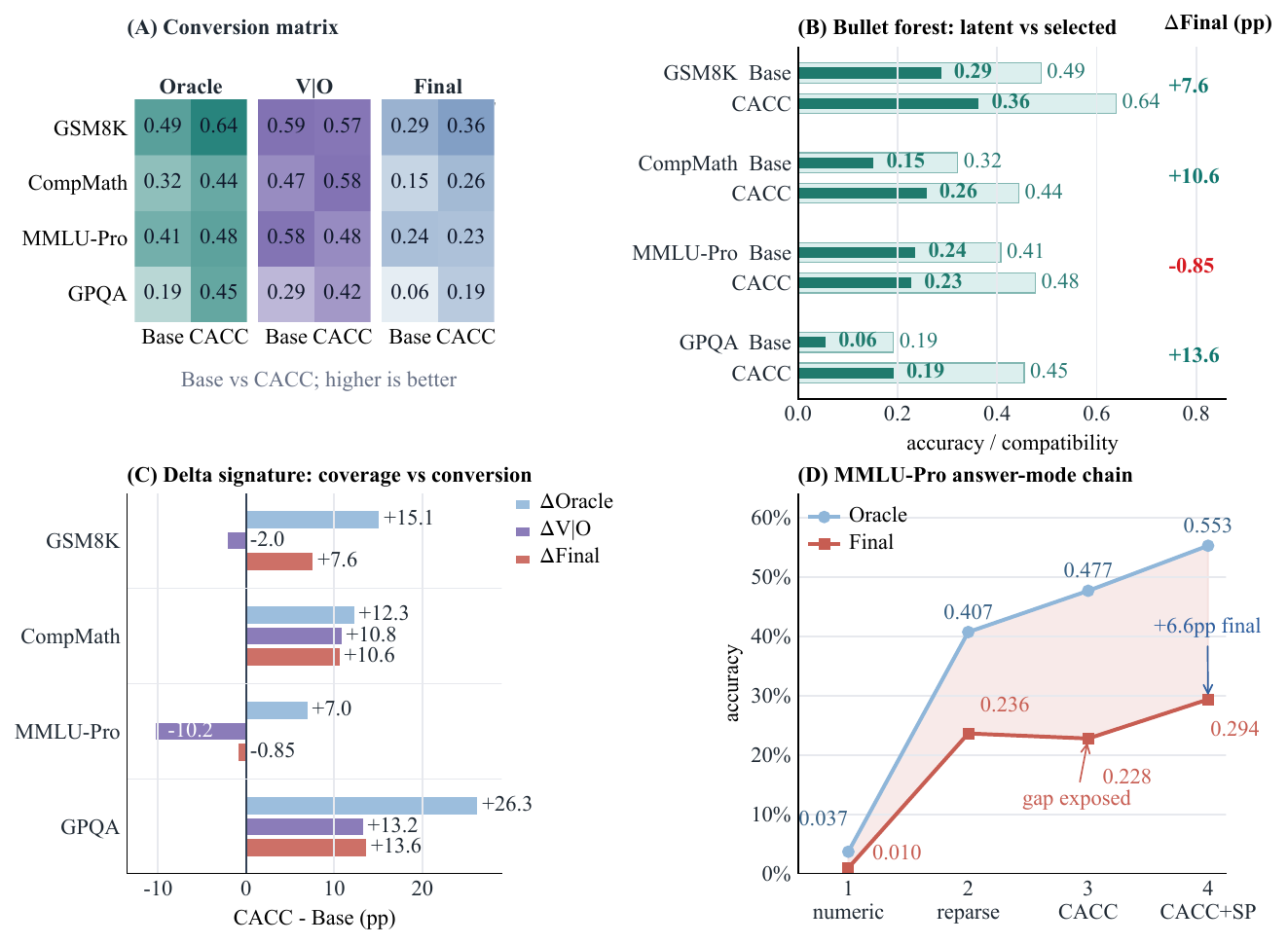}
\caption{Main results across reasoning regimes. Panel A: Base/CACC oracle, V|O, and final. Panel B: latent coverage upper bound and selected final accuracy with \CACC{}-vs-Base delta. Panel C: $\Delta$oracle, $\Delta$V|O, and $\Delta$final separating coverage, conversion, and final. Panel D: four-step MMLU-Pro mechanism chain. Exact numbers are in Table~\ref{tab:main-results} and Appendix Tables~\ref{tab:decomposition},~\ref{tab:table4}.}
\label{fig:main-results}
\end{figure}

Table~\ref{tab:main-results} reports unified method rows for Base, \SPP{}, \CACC{}, and \CACC{}+\SPP{} across four datasets. \SPP{} denotes proposer-side strengthening with a Qwen3-8B proposer, a different compute regime from the Qwen3-1.7B generator used by Base and \CACC{}. For numeric tasks, Base is the raw generate-then-rerank pool. For MMLU-Pro, Base is a CCR-only legacy pool scored under benchmark-locked MCQ reparse (see Appendix~\ref{app:mmlu-provenance}). Final $=$ Oracle $\times$ V|O, and Ratio is Final divided by the Base-row Final on the same dataset. The MMLU-Pro no-\AMC{} archived numeric provenance (oracle $0.0369$, final $0.0096$) is used only for AMC necessity; see Appendix~\ref{app:mmlu-provenance}.

\begin{table}[t]
\centering
\footnotesize
\caption{Candidate-interface results. $\Delta$Final and Ratio are computed against the Base row within the same dataset. $^\dagger$ uses a Qwen3-8B proposer (different compute regime from the Qwen3-1.7B generator). Paired tests for \CACC{} vs Base are in Appendix Table~\ref{tab:app-paired-details}.}
\label{tab:main-results}
\setlength{\tabcolsep}{3pt}
\renewcommand{\arraystretch}{1.0}
\begin{tabular}{@{}l l l c c c c c@{}}
\toprule
Regime & Dataset & Method & Oracle & V|O & Final & $\Delta$Final & Ratio \\
\midrule
Numeric & GSM8K & Base & 0.4882 & 0.5885 & 0.2873 & ref. & 1.00$\times$ \\
Numeric & GSM8K & \SPP{}$^\dagger$ & 0.6471 & 0.5714 & 0.3698 & +8.25pp & 1.29$\times$ \\
Numeric & GSM8K & \CACC{} & 0.6391 & 0.5682 & 0.3632 & +7.59pp & 1.26$\times$ \\
Numeric & GSM8K & \CACC{}+\SPP{}$^\dagger$ & 0.6962 & 0.6017 & 0.4189 & +13.16pp & 1.46$\times$ \\
Numeric & CompMath & Base & 0.3210 & 0.4732 & 0.1519 & ref. & 1.00$\times$ \\
Numeric & CompMath & \SPP{}$^\dagger$ & 0.4616 & 0.6014 & 0.2776 & +12.57pp & 1.83$\times$ \\
Numeric & CompMath & \CACC{} & 0.4436 & 0.5814 & 0.2579 & +10.60pp & 1.70$\times$ \\
Numeric & CompMath & \CACC{}+\SPP{}$^\dagger$ & 0.5024 & 0.6173 & 0.3101 & +15.82pp & 2.04$\times$ \\
MCQ & MMLU-Pro & Base & 0.4072 & 0.5800 & 0.2362 & ref. & 1.00$\times$ \\
MCQ & MMLU-Pro & \SPP{}$^\dagger$ & 0.4921 & 0.5411 & 0.2663 & +3.01pp & 1.13$\times$ \\
MCQ & MMLU-Pro & \CACC{} & 0.4767 & 0.4777 & 0.2277 & -0.85pp & 0.96$\times$ \\
MCQ & MMLU-Pro & \CACC{}+\SPP{}$^\dagger$ & 0.5527 & 0.5312 & 0.2936 & +5.74pp & 1.24$\times$ \\
MCQ & GPQA & Base & 0.1919 & 0.2897 & 0.0556 & ref. & 1.00$\times$ \\
MCQ & GPQA & \SPP{}$^\dagger$ & 0.4724 & 0.4197 & 0.1983 & +14.27pp & 3.57$\times$ \\
MCQ & GPQA & \CACC{} & 0.4545 & 0.4222 & 0.1919 & +13.63pp & 3.45$\times$ \\
MCQ & GPQA & \CACC{}+\SPP{}$^\dagger$ & 0.4899 & 0.5258 & 0.2576 & +20.20pp & 4.63$\times$ \\
\bottomrule
\end{tabular}
\end{table}

\paragraph{Numeric reasoning and coverage--conversion decomposition.}
\label{sec:rq1-numeric}
The dominant failure modes on numeric open-ended tasks are fragments, partial solutions, and missing-final answers. The numeric block of Table~\ref{tab:main-results} together with Panels~A--B of Figure~\ref{fig:main-results} shows: GSM8K clean oracle $0.4882\to 0.6391$ and final $0.2873\to 0.3632$; \texttt{competition\_math:numeric} oracle $0.3210\to 0.4436$ and final $0.1519\to 0.2579$. Under a fixed selector, \CCR{}'s repair of incomplete reasoning and missing-final answers improves both candidate coverage and final selection. However, coverage, V|O, and final do not move in lockstep (Table~\ref{tab:decomposition}): the answer-mode interface can raise oracle and final together. Once readout is aligned, micro-ablations yield small marginal gains. a \emph{diverse pool $\to$ \CACC{}} transition can lower final when oracle rises but V|O drops, and adding hygiene then recovers final via V|O; on GSM8K full, SP $\to$ Power Sampling (PS)$\times$\CACC{} shows raw oracle $+10.84$pp can still come with V|O and final drops. We therefore report the coverage--conversion decomposition rather than a single oracle/final indicator. In the numeric regime, \CACC{}'s gains transfer cleanly because the numeric answer mode imposes a lightweight readout contract. \CCR{}'s repair of fragments and missing-final answers maps directly to selectable coverage. The MMLU-Pro micro-ablation in Table~\ref{tab:decomposition} (CCR+\AMC{}$\to$\CACC{}, $\Delta F=+0.07$\,pp) shows a small marginal gain from hygiene in this setting.

\begin{table}[t]
\centering
\footnotesize
\caption{Coverage--selection decomposition on audit rows (outside Table~\ref{tab:main-results}). $O,V,F$ denote oracle, V|O, and final; $b/a$ denote before/after; $\Delta$ in percentage points.}
\label{tab:decomposition}
\setlength{\tabcolsep}{2.4pt}
\renewcommand{\arraystretch}{1.0}
\begin{tabular}{@{}l l c c c c c c r r r@{}}
\toprule
Unit & Comparison & $O_b$ & $O_a$ & $V_b$ & $V_a$ & $F_b$ & $F_a$ & $\Delta O$ & $\Delta V$ & $\Delta F$ \\
\midrule
MMLU-Pro & no-\AMC{} numeric $\rightarrow$ \CACC{} & 0.0369 & 0.4767 & 0.2590 & 0.4777 & 0.0096 & 0.2277 & +43.98 & +21.87 & +21.81 \\
MMLU-Pro & CCR+\AMC{} $\rightarrow$ \CACC{} & 0.4775 & 0.4767 & 0.4754 & 0.4777 & 0.2270 & 0.2277 & -0.08 & +0.23 & +0.07 \\
MMLU-512 & diverse $\rightarrow$ \CACC{} & 0.4609 & 0.5547 & 0.4492 & 0.3592 & 0.2070 & 0.1992 & +9.38 & -9.00 & -0.78 \\
MMLU-512 & \CACC{} $\rightarrow$ +hygiene & 0.5547 & 0.5547 & 0.3592 & 0.4049 & 0.1992 & 0.2246 & +0.00 & +4.58 & +2.54 \\
GSM8K full & SP $\rightarrow$ PS x \CACC{} & 0.6111 & 0.7195 & 0.5943 & 0.4700 & 0.3632 & 0.3381 & +10.84 & -12.43 & -2.50 \\
\bottomrule
\end{tabular}
\end{table}

\paragraph{MCQ readout chain and second-family closure.}
\label{sec:rq3-mcq}
What matters for MCQ is not the sheer number of candidates but the supply of answer-mode-compatible candidates that the fixed selector can consume. MCQ tasks impose stronger readout constraints on the candidate interface (Figure~\ref{fig:main-results} Panel D). Under archived numeric scoring, the legacy/no-\AMC{} candidate interface yields a final of only $0.0096$. Switching to benchmark-locked choice-letter readout, the same-family CCR-only baseline recovers oracle/final to $0.4072/0.2362$ (Appendix~\ref{app:mmlu-provenance}). \CACC{} further raises oracle to $0.4767$ but holds final at $0.2277$, exposing the coverage--conversion gap, and \CACC{}+\SPP{} pushes final to $0.2936$ in the selected regime.

GPQA Diamond exhibits the same direction with a more pronounced effect. Under the fixed-verifier interface, the Base final is only $0.0556$ partly because answer-mode alignment is missing and many candidates are rejected by the readout as invalid (the predicted consequence of the hidden interface tax). \CACC{}-only raises final to $0.1919$ ($+13.6$pp absolute gain), and \CACC{}+\SPP{} reaches $0.2576$ ($+20.2$pp). Second-family paired closure further supports recoverable compatibility: on MMLU-Pro confirm-512, \CACC{}$\to$+hygiene moves $0.1992\to 0.2246$ (McNemar $p=0.0072$); on GPQA-128, $0.2031\to 0.2656$ ($p=0.0215$). The Mistral-7B Smoke rows trend in the same direction (Table~\ref{tab:table4} and Appendix Table~\ref{tab:app-paired-details}). The MMLU-Pro coverage--conversion gap is therefore not a failure mode of \CACC{} but an explicit revelation. Under a fixed verifier and a strict choice-letter readout, raising oracle without aligning the answer-mode contract converts only after proposer-side strengthening reshapes the candidate distribution toward verifier-friendly options. GPQA Diamond shows the same pattern at higher amplitude precisely because its base is more strongly limited by readout invalidity, which is also why second-family hygiene (rather than a transplanted rulebook) recovers selectable coverage when the proposer changes.

\paragraph{Layer-aware positioning and gap-narrowing interactions.}
\label{sec:rq6-interaction}
\CACC{} should be understood relative to external enhancements at the system-layer level. Self-Refine improves trajectories, ToT/search expands the reasoning space, self-consistency expands the candidate source, FTTT applies fine-tuning at test time, and verifiers/rerankers change selection. \CACC{} improves the candidate-interface distribution before any of these sources reaches the selector (per-method compute/budget/layer differences are listed in Appendix Table~\ref{tab:app-external-provenance}). The coverage--conversion gap can be narrowed jointly across these layers (Table~\ref{tab:table4} and Appendix Figure~\ref{fig:mechanism}): on the proposer side, \CACC{}+\SPP{} raises final from $0.2277/0.1919$ to $0.2936/0.2576$ on MMLU-Pro and GPQA respectively. Layering conversion-aware rescue further improves selection-side conversion (MMLU-Pro $0.2936\to 0.3320$, $p=3.43\times 10^{-67}$; GPQA $0.2576\to 0.2929$, $p=0.0923$). Composability across layers is constructive but not additive: \CACC{}+\SPP{} narrows the gap from the source side, conversion-aware rescue narrows it from the selection side, and the Power Sampling$\times$\CACC{} diagnostic in Table~\ref{tab:table4} shows that distribution shaping at the wrong layer can widen rather than narrow the gap. Layer matching (not the number of stacked enhancements) is what predicts whether a given composition converts oracle expansion into final accuracy under a fixed selector.

\begin{table}[t]
\centering
\footnotesize
\caption{Ablation, recovery, and interaction evidence. Reports oracle, V|O, and final before$\rightarrow$after with the final delta; sample sizes, wins/losses, $p$-values, scope, and artifact statuses are in Appendix Tables~\ref{tab:app-paired-details},~\ref{tab:app-artifacts}, and~\ref{tab:app-external-provenance}.}
\label{tab:table4}
\setlength{\tabcolsep}{1.5pt}
\renewcommand{\arraystretch}{1.0}
\begin{tabular}{@{}l l l c c c c@{}}
\toprule
Type & Dataset & Method / transition & Oracle & V|O & Final & $\Delta F$ \\
\midrule
\textsc{Readout} & MMLU-Pro & numeric scoring $\rightarrow$ MCQ reparse & 0.0369$\rightarrow$0.4072 & 0.2590$\rightarrow$0.5800 & 0.0096$\rightarrow$0.2362 & +22.66pp \\
\textsc{AMC} & MMLU-Pro & MCQ reparse $\rightarrow$ \CACC{} & 0.4072$\rightarrow$0.4767 & 0.5800$\rightarrow$0.4777 & 0.2362$\rightarrow$0.2277 & -0.85pp \\
\textsc{SP} & MMLU-Pro & \CACC{} $\rightarrow$ \CACC{}+\SPP{} & 0.4767$\rightarrow$0.5527 & 0.4777$\rightarrow$0.5312 & 0.2277$\rightarrow$0.2936 & +6.59pp \\
\textsc{SP} & GPQA & \CACC{} $\rightarrow$ \CACC{}+\SPP{} & 0.4545$\rightarrow$0.4899 & 0.4222$\rightarrow$0.5258 & 0.1919$\rightarrow$0.2576 & +6.57pp \\
\textsc{Recovery} & MMLU-Pro & \CACC{} $\rightarrow$ +hygiene & 0.5547$\rightarrow$0.5547 & 0.3592$\rightarrow$0.4049 & 0.1992$\rightarrow$0.2246 & +2.54pp \\
\textsc{Recovery} & GPQA & \CACC{} $\rightarrow$ +hygiene & 0.5127$\rightarrow$0.4931 & 0.3961$\rightarrow$0.5386 & 0.2031$\rightarrow$0.2656 & +6.25pp \\
\textsc{Rescue} & MMLU-Pro & \CACC{}+\SPP{} $\rightarrow$ +rescue & 0.5527$\rightarrow$0.5527 & 0.5312$\rightarrow$0.6007 & 0.2936$\rightarrow$0.3320 & +3.84pp \\
\textsc{Rescue} & GPQA & \CACC{}+\SPP{} $\rightarrow$ +rescue & 0.4899$\rightarrow$0.4899 & 0.5258$\rightarrow$0.5979 & 0.2576$\rightarrow$0.2929 & +3.53pp \\
\textsc{Diagnostic} & GSM8K & \SPP{} $\rightarrow$ PS $\times$ \CACC{} & 0.6111$\rightarrow$0.7195 & 0.5943$\rightarrow$0.4700 & 0.3632$\rightarrow$0.3381 & -2.50pp \\
\textsc{Smoke} & GSM8K & Mistral diverse $\rightarrow$ selector-cal & 0.3896$\rightarrow$0.4179 & 0.5413$\rightarrow$0.5422 & 0.2109$\rightarrow$0.2266 & +1.56pp \\
\textsc{Smoke} & MMLU-Pro & Mistral diverse $\rightarrow$ filter+hygiene & 0.4129$\rightarrow$0.4367 & 0.4541$\rightarrow$0.4829 & 0.1875$\rightarrow$0.2109 & +2.34pp \\
\bottomrule
\end{tabular}
\end{table}

\subsection{Additional Controls and Prospective Validation}
\label{sec:controls}
Table~\ref{tab:control-summary} adds the key controls; protocols and paired statistics are in Appendix~\ref{app:controls}.

\textbf{Trained verifiers and selector families.}
With non-evaluation supervision and canonical choice rendering, all three target-verifier seeds improve on GPQA and held-out MMLU-Pro. On frozen GPQA pools, majority voting, 8B option-logprob, and 1.7B/32B judges give gains of $3.03$, $10.61$, $13.13$, and $16.67$\,pp. MMLU effects depend on selector/readout and scope. Mistral generation improves on all five seeds, with a mean gain of $1.72$\,pp.

\textbf{Constrained decoding and format.}
The fresh four-cell control shares Qwen3-1.7B stochastic prefixes and a target-specific canonical selector. \CACC{} improves final accuracy over Base by $6.64$\,pp on MMLU confirm-512. CD exceeds \CACC{} by $6.57$\,pp on GPQA and $0.98$\,pp on MMLU confirm-512. CD+\CACC{} adds coverage without improving final accuracy over CD. Content-preserving final-marker changes flip $9.89\%$ of eligible MMLU selections, with little change in aggregate accuracy.

\begin{table}[t]
\centering\footnotesize
\caption{Additional controls. Final accuracies are percentages; gains are percentage points. Frozen-pool rows reuse the original candidates. Target-verifier rows are three-seed means; Mistral is a five-seed mean. CD uses fresh paired generation and a common canonical selector, with its own baseline.}
\label{tab:control-summary}
\setlength{\tabcolsep}{3pt}
\begin{tabular}{@{}l l r r r r@{}}
\toprule Control & Evaluation & $N$ & Base & \CACC{} & $\Delta F$\\\midrule
Target verifier, canonical & GPQA & 198 & 16.16 & 24.92 & +8.75\\
Target verifier, canonical & MMLU held-out & 11,520 & 26.40 & 27.84 & +1.44\\
32B candidate judge & GPQA & 198 & 6.57 & 23.23 & +16.67\\
Mistral generator & MMLU confirm & 512 & 14.22 & 15.94 & +1.72\\
Prospective ARC reuse & ARC-Challenge & 256 & 69.92 & 70.70 & +0.78\\\midrule
Fresh generation & Evaluation & $N$ & \CACC{} & CD & CD+\CACC{}\\
Paired four-cell control & GPQA & 198 & 23.23 & 29.80 & 28.79\\
Paired four-cell control & MMLU confirm & 512 & 29.30 & 30.27 & 28.12\\
\bottomrule\end{tabular}
\end{table}

\textbf{MMLU conversion and new-benchmark reuse.}
The MMLU-Pro majority-vote analysis shows limited conversion of newly added coverage and losses of previously available correct candidates. On ARC-Challenge, reusing the unchanged rules increases oracle coverage with little change in final accuracy (Appendix~\ref{sec:arc-prospective}).

\paragraph{Qualitative cases.}
The error taxonomy in Appendix Table~\ref{tab:app-error-taxonomy} aligns with concrete audit examples (Appendix Table~\ref{tab:app-audit-examples}). On \texttt{gsm8k-test-00462}, the original Qwen3-1.7B candidate stops at a fragmented arithmetic step. \CCR{} repairs the partial derivation into a boxed final answer that the verifier then promotes. On \texttt{mmlu-pro-test-08795}, the legacy/no-\AMC{} numeric view misses a choice-form solution, while \CACC{}+\SPP{} contributes an answer-mode-aligned candidate that recovers the final answer A. On \texttt{gpqa-diamond-train-recNFJjE5PPTqVJGv}, the wrongly selected candidate carries scaffold/instruction residue, illustrating that hygiene (rather than further sampling) determines whether the verifier latches onto a correct option. Together these cases trace a one-to-one mapping from interface failure modes to the oracle / V|O / final movements reported above.

\section{Discussion}

\textbf{Composability and deployment.} Because the offline rulebook is frozen before online inference, the same rulebook composes with proposer-side strengthening or conversion-aware rescue without changing the verifier. For deployment, an interface audit is a low-cost first step before investing in further sampling or feedback rounds. The MMLU-Pro readout chain and GPQA Diamond results show that interface alignment and candidate repair can convert latent oracle coverage into final accuracy.

\textbf{Limitations.} The current study targets numeric and MCQ answer modes with benchmark-locked parsers. Extending to free-form, code-generation, and multi-turn tasks requires generalizing the answer-mode registry beyond fixed parser contracts. The offline rulebook is manually calibrated per benchmark, and learning the routing policy end-to-end from selector feedback would reduce setup cost. Additional experiments show that the effect of \CACC{} depends on the selector, readout, and evaluation scope; the accuracy decline in the original MMLU-Pro setting illustrates this limitation. The current ablation covers CCR+AMC, CCR-only, and hygiene-only transitions (Tables~\ref{tab:decomposition}--\ref{tab:table4}). Finer parser-only ablations are left to future work.

\section{Conclusion}

This paper identified the hidden interface tax in multi-candidate generate-then-rerank reasoning, where correct candidates appear in the pool but cannot be consumed by the fixed selector. We formalized this as the coverage--conversion decomposition, introduced selectable coverage as the operative upper bound, and proposed \CACC{} to close the gap through offline-calibrated fragment repair, answer-mode alignment, and candidate hygiene. On numeric reasoning, \CACC{} raises both oracle and final accuracy under a fixed verifier. On GPQA Diamond, \CACC{} alone yields $+13.6$\,pp and composition with proposer-side strengthening reaches $+20.2$\,pp, demonstrating that interface-level intervention can convert latent oracle coverage into final accuracy without retraining any component.

\begin{ack}
This research received support from the National Natural Science Foundation of China (grants 62272050 and 62302048); the Guangdong Provincial Department of Education's 2023--2024 support for the Guangdong Key Lab of AI and Multi-modal Data Processing at Beijing Normal-Hong Kong Baptist University (BNBU), Zhuhai; the Institute of Artificial Intelligence and Future Networks (BNU-Zhuhai); the Engineering Center of AI and Future Education of the Guangdong Provincial Department of Science and Technology, China; and the Zhuhai Science-Tech Innovation Bureau (grant 2320004002772). We also acknowledge computing support from the Interdisciplinary Intelligence SuperComputer Center of Beijing Normal University (Zhuhai).

\textbf{Competing interests:} The authors declare no competing interests.
\end{ack}

\bibliographystyle{abbrvnat}

\bibliography{references}

@article{christiano2017preferences,
  title={Deep reinforcement learning from human preferences},
  author={Christiano, Paul F and Leike, Jan and Brown, Tom and Martic, Miljan and Legg, Shane and Amodei, Dario},
  journal={Advances in neural information processing systems},
  volume={30},
  year={2017}
}

@article{cobbe2021training,
  title={Training verifiers to solve math word problems},
  author={Cobbe, Karl and Kosaraju, Vineet and Bavarian, Mohammad and Chen, Mark and Jun, Heewoo and Kaiser, Lukasz and Plappert, Matthias and Tworek, Jerry and Hilton, Jacob and Nakano, Reiichiro and others},
  journal={arXiv preprint arXiv:2110.14168},
  year={2021},
  url={https://arxiv.org/abs/2110.14168}
}

@article{hendrycks2021math,
  title={Measuring mathematical problem solving with the math dataset},
  author={Hendrycks, Dan and Burns, Collin and Kadavath, Saurav and Arora, Akul and Basart, Steven and Tang, Eric and Song, Dawn and Steinhardt, Jacob},
  journal={arXiv preprint arXiv:2103.03874},
  year={2021}
}

@article{hendrycks2021mmlu,
  title={Measuring massive multitask language understanding},
  author={Hendrycks, Dan and Burns, Collin and Basart, Steven and Zou, Andy and Mazeika, Mantas and Song, Dawn and Steinhardt, Jacob},
  journal={arXiv preprint arXiv:2009.03300},
  year={2020}
}

@article{ouyang2022training,
  title={Training language models to follow instructions with human feedback},
  author={Ouyang, Long and Wu, Jeffrey and Jiang, Xu and Almeida, Diogo and Wainwright, Carroll and Mishkin, Pamela and Zhang, Chong and Agarwal, Sandhini and Slama, Katarina and Ray, Alex and others},
  journal={Advances in neural information processing systems},
  volume={35},
  pages={27730--27744},
  year={2022}
}

@article{uesato2022solving,
  title={Solving math word problems with process-and outcome-based feedback},
  author={Uesato, Jonathan and Kushman, Nate and Kumar, Ramana and Song, Francis and Siegel, Noah and Wang, Lisa and Creswell, Antonia and Irving, Geoffrey and Higgins, Irina},
  journal={arXiv preprint arXiv:2211.14275},
  year={2022}
}

@article{wei2022cot,
  title={Chain-of-thought prompting elicits reasoning in large language models},
  author={Wei, Jason and Wang, Xuezhi and Schuurmans, Dale and Bosma, Maarten and Xia, Fei and Chi, Ed and Le, Quoc V and Zhou, Denny and others},
  journal={Advances in neural information processing systems},
  volume={35},
  pages={24824--24837},
  year={2022}
}

@article{kojima2022zeroshot,
  title={Large language models are zero-shot reasoners},
  author={Kojima, Takeshi and Gu, Shixiang Shane and Reid, Machel and Matsuo, Yutaka and Iwasawa, Yusuke},
  journal={Advances in neural information processing systems},
  volume={35},
  pages={22199--22213},
  year={2022}
}

@article{bai2022constitutional,
  title={Constitutional ai: Harmlessness from ai feedback},
  author={Bai, Yuntao and Kadavath, Saurav and Kundu, Sandipan and Askell, Amanda and Kernion, Jackson and Jones, Andy and Chen, Anna and Goldie, Anna and Mirhoseini, Azalia and McKinnon, Cameron and others},
  journal={arXiv preprint arXiv:2212.08073},
  year={2022},
  url={https://arxiv.org/abs/2212.08073}
}

@article{lewkowycz2022minerva,
  title={Solving quantitative reasoning problems with language models},
  author={Lewkowycz, Aitor and Andreassen, Anders and Dohan, David and Dyer, Ethan and Michalewski, Henryk and Ramasesh, Vinay and Slone, Ambrose and Anil, Cem and Schlag, Imanol and Gutman-Solo, Theo and others},
  journal={Advances in neural information processing systems},
  volume={35},
  pages={3843--3857},
  year={2022}
}

@inproceedings{wang2023selfconsistency,
  title={Self-consistency improves chain of thought reasoning in language models},
  author={Wang, Xuezhi and Wei, Jason and Schuurmans, Dale and Le, Quoc and Chi, Ed and Narang, Sharan and Chowdhery, Aakanksha and Zhou, Denny},
  booktitle={International Conference on Learning Representations},
  year={2023},
  url={https://arxiv.org/abs/2203.11171}
}

@inproceedings{gao2023pal,
  title={Pal: Program-aided language models},
  author={Gao, Luyu and Madaan, Aman and Zhou, Shuyan and Alon, Uri and Liu, Pengfei and Yang, Yiming and Callan, Jamie and Neubig, Graham},
  booktitle={International conference on machine learning},
  pages={10764--10799},
  year={2023},
  organization={PMLR}
}

@article{yao2023react,
  title={React: Synergizing reasoning and acting in language models},
  author={Yao, Shunyu and Zhao, Jeffrey and Yu, Dian and Du, Nan and Shafran, Izhak and Narasimhan, Karthik and Cao, Yuan},
  journal={arXiv preprint arXiv:2210.03629},
  year={2022},
  url={https://arxiv.org/abs/2210.03629}
}

@article{yao2023tot,
  title={Tree of thoughts: Deliberate problem solving with large language models},
  author={Yao, Shunyu and Yu, Dian and Zhao, Jeffrey and Shafran, Izhak and Griffiths, Tom and Cao, Yuan and Narasimhan, Karthik},
  journal={Advances in neural information processing systems},
  volume={36},
  pages={11809--11822},
  year={2023}
}

@article{madaan2023selfrefine,
  title={Self-refine: Iterative refinement with self-feedback},
  author={Madaan, Aman and Tandon, Niket and Gupta, Prakhar and Hallinan, Skyler and Gao, Luyu and Wiegreffe, Sarah and Alon, Uri and Dziri, Nouha and Prabhumoye, Shrimai and Yang, Yiming and others},
  journal={arXiv preprint arXiv:2303.17651},
  year={2023},
  url={https://arxiv.org/abs/2303.17651}
}

@article{shinn2023reflexion,
  title={Reflexion: Language agents with verbal reinforcement learning},
  author={Shinn, Noah and Cassano, Federico and Gopinath, Ashwin and Narasimhan, Karthik and Yao, Shunyu},
  journal={Advances in neural information processing systems},
  volume={36},
  pages={8634--8652},
  year={2023}
}

@article{rafailov2023dpo,
  title={Direct preference optimization: Your language model is secretly a reward model},
  author={Rafailov, Rafael and Sharma, Archit and Mitchell, Eric and Manning, Christopher D and Ermon, Stefano and Finn, Chelsea},
  journal={Advances in neural information processing systems},
  volume={36},
  pages={53728--53741},
  year={2023}
}

@article{rein2023gpqa,
  title={Gpqa: A graduate-level google-proof q\&a benchmark},
  author={Rein, David and Hou, Betty Li and Stickland, Asa Cooper and Petty, Jackson and Pang, Richard Yuanzhe and Dirani, Julien and Michael, Julian and Bowman, Samuel R},
  journal={arXiv preprint arXiv:2311.12022},
  year={2023}
}

@article{zheng2023judging,
  title={Judging llm-as-a-judge with mt-bench and chatbot arena},
  author={Zheng, Lianmin and Chiang, Wei-Lin and Sheng, Ying and Zhuang, Siyuan and Wu, Zhanghao and Zhuang, Yonghao and Lin, Zi and Li, Zhuohan and Li, Dacheng and Xing, Eric and others},
  journal={Advances in neural information processing systems},
  volume={36},
  pages={46595--46623},
  year={2023}
}

@misc{jiang2023mistral,
      title={Mistral 7B}, 
      author={Albert Q. Jiang and Alexandre Sablayrolles and Arthur Mensch and Chris Bamford and Devendra Singh Chaplot and Diego de las Casas and Florian Bressand and Gianna Lengyel and Guillaume Lample and Lucile Saulnier and Lélio Renard Lavaud and Marie-Anne Lachaux and Pierre Stock and Teven Le Scao and Thibaut Lavril and Thomas Wang and Timothée Lacroix and William El Sayed},
      year={2023},
      eprint={2310.06825},
      archivePrefix={arXiv},
      primaryClass={cs.CL},
      url={https://arxiv.org/abs/2310.06825}, 
}

@inproceedings{lightman2024verify,
  title={Let's verify step by step},
  author={Lightman, Hunter and Kosaraju, Vineet and Burda, Yuri and Edwards, Harrison and Baker, Bowen and Lee, Teddy and Leike, Jan and Schulman, John and Sutskever, Ilya and Cobbe, Karl},
  booktitle={International Conference on Learning Representations},
  year={2024},
  url={https://proceedings.iclr.cc/paper_files/paper/2024/hash/aca97732e30bcf1303bc22ac3924fd16-Abstract-Conference.html}
}

@article{miao2024selfcheck,
  title={Selfcheck: Using llms to zero-shot check their own step-by-step reasoning},
  author={Miao, Ning and Teh, Yee Whye and Rainforth, Tom},
  journal={arXiv preprint arXiv:2308.00436},
  year={2023},
  url={https://arxiv.org/abs/2308.00436}
}

@inproceedings{wang2024mathshepherd,
  title={Math-Shepherd: Verify and Reinforce {LLMs} Step-by-step without Human Annotations},
  author={Wang, Peiyi and Li, Lei and Shao, Zhihong and Xu, Runxin and Dai, Damai and Li, Yifei and Chen, Deli and Wu, Yu and Sui, Zhifang},
  booktitle={Proceedings of the 62nd Annual Meeting of the Association for Computational Linguistics (Volume 1: Long Papers)},
  pages={9426--9439},
  year={2024},
  url={https://aclanthology.org/2024.acl-long.510/}
}

@article{kumar2024selfcorrect,
  title={Training language models to self-correct via reinforcement learning},
  author={Kumar, Aviral and Zhuang, Vincent and Agarwal, Rishabh and Su, Yi and Co-Reyes, John D and Singh, Avi and Baumli, Kate and Iqbal, Shariq and Bishop, Colton and Roelofs, Rebecca and others},
  journal={arXiv preprint arXiv:2409.12917},
  year={2024},
  url={https://arxiv.org/abs/2409.12917}
}

@inproceedings{li2024chaincode,
  title={Chain of code: Reasoning with a language model-augmented code emulator},
  author={Li, Chengshu and Liang, Jacky and Zeng, Andy and Chen, Xinyun and Hausman, Karol and Sadigh, Dorsa and Levine, Sergey and Fei-Fei, Li and Xia, Fei and Ichter, Brian},
  booktitle={International Conference on Machine Learning},
  year={2024},
  url={https://arxiv.org/abs/2312.04474}
}

@article{zhou2024lats,
  title={Language agent tree search unifies reasoning acting and planning in language models},
  author={Zhou, Andy and Yan, Kai and Shlapentokh-Rothman, Michal and Wang, Haohan and Wang, Yu-Xiong},
  journal={arXiv preprint arXiv:2310.04406},
  year={2023},
  url={https://arxiv.org/abs/2310.04406}
}

@inproceedings{yuan2024selfrewarding,
  title={Self-rewarding language models},
  author={Yuan, Weizhe and Pang, Richard Yuanzhe and Cho, Kyunghyun and Li, Xian and Sukhbaatar, Sainbayar and Xu, Jing and Weston, Jason},
  booktitle={International Conference on Machine Learning},
  year={2024},
  url={https://arxiv.org/abs/2401.10020}
}

@article{zelikman2024quietstar,
  title={Quiet-star: Language models can teach themselves to think before speaking},
  author={Zelikman, Eric and Harik, Georges and Shao, Yijia and Jayasiri, Varuna and Haber, Nick and Goodman, Noah D},
  journal={arXiv preprint arXiv:2403.09629},
  year={2024},
  url={https://arxiv.org/abs/2403.09629}
}

@article{zhou2024selfdiscover,
  title={Self-discover: Large language models self-compose reasoning structures},
  author={Zhou, Pei and Pujara, Jay and Ren, Xiang and Chen, Xinyun and Cheng, Heng-Tze and Le, Quoc V and Zhou, Denny and Mishra, Swaroop and Zheng, Huaixiu S and others},
  journal={Advances in Neural Information Processing Systems},
  volume={37},
  pages={126032--126058},
  year={2024}
}

@article{setlur2024rewarding,
  title={Rewarding progress: Scaling automated process verifiers for llm reasoning},
  author={Setlur, Amrith and Nagpal, Chirag and Fisch, Adam and Geng, Xinyang and Eisenstein, Jacob and Agarwal, Rishabh and Agarwal, Alekh and Berant, Jonathan and Kumar, Aviral},
  journal={arXiv preprint arXiv:2410.08146},
  year={2024}
}

@inproceedings{lambert2024rewardbench,
  title={Rewardbench: Evaluating reward models for language modeling},
  author={Lambert, Nathan and Pyatkin, Valentina and Morrison, Jacob and Miranda, LJ and Lin, Bill Yuchen and Chandu, Khyathi and Dziri, Nouha and Kumar, Sachin and Zick, Tom and Choi, Yejin and others},
  booktitle={Findings of the Association for Computational Linguistics: NAACL 2025},
  pages={1755--1797},
  year={2025}
}

@article{wang2024mmlupro,
  title={Mmlu-pro: A more robust and challenging multi-task language understanding benchmark},
  author={Wang, Yubo and Ma, Xueguang and Zhang, Ge and Ni, Yuansheng and Chandra, Abhranil and Guo, Shiguang and Ren, Weiming and Arulraj, Aaran and He, Xuan and Jiang, Ziyan and others},
  journal={Advances in Neural Information Processing Systems},
  volume={37},
  pages={95266--95290},
  year={2024}
}

@article{chen2024livebench,
  title={Livebench: A challenging, contamination-free llm benchmark},
  author={White, Colin and Dooley, Samuel and Roberts, Manley and Pal, Arka and Feuer, Ben and Jain, Siddhartha and Shwartz-Ziv, Ravid and Jain, Neel and Saifullah, Khalid and Naidu, Siddartha and others},
  journal={arXiv preprint arXiv:2406.19314},
  year={2024},
  url={https://arxiv.org/abs/2406.19314}
}

@article{brown2024monkeys,
  title={Large language monkeys: Scaling inference compute with repeated sampling},
  author={Brown, Bradley and Juravsky, Jordan and Ehrlich, Ryan and Clark, Ronald and Le, Quoc V and R{\'e}, Christopher and Mirhoseini, Azalia},
  journal={arXiv preprint arXiv:2407.21787},
  year={2024}
}

@article{qwen2024qwen25,
  title={Qwen2. 5-coder technical report},
  author={Hui, Binyuan and Yang, Jian and Cui, Zeyu and Yang, Jiaxi and Liu, Dayiheng and Zhang, Lei and Liu, Tianyu and Zhang, Jiajun and Yu, Bowen and Lu, Keming and others},
  journal={arXiv preprint arXiv:2409.12186},
  year={2024}
}

@article{yang2024qwen25math,
  title={Qwen2. 5-math technical report: Toward mathematical expert model via self-improvement},
  author={Yang, An and Zhang, Beichen and Hui, Binyuan and Gao, Bofei and Yu, Bowen and Li, Chengpeng and Liu, Dayiheng and Tu, Jianhong and Zhou, Jingren and Lin, Junyang and others},
  journal={arXiv preprint arXiv:2409.12122},
  year={2024}
}

@article{dubey2024llama3,
  title={The llama 3 herd of models},
  author={Grattafiori, Aaron and Dubey, Abhimanyu and Jauhri, Abhinav and Pandey, Abhinav and Kadian, Abhishek and Al-Dahle, Ahmad and Letman, Aiesha and Mathur, Akhil and Schelten, Alan and Vaughan, Alex and others},
  journal={arXiv preprint arXiv:2407.21783},
  year={2024}
}

@article{team2024gemma,
  title={Gemma: Open models based on gemini research and technology},
  author={Team, Gemma and Mesnard, Thomas and Hardin, Cassidy and Dadashi, Robert and Bhupatiraju, Surya and Pathak, Shreya and Sifre, Laurent and Rivi{\`e}re, Morgane and Kale, Mihir Sanjay and Love, Juliette and others},
  journal={arXiv preprint arXiv:2403.08295},
  year={2024}
}

@article{abdin2024phi3,
  title={Phi-4 technical report},
  author={Abdin, Marah and Aneja, Jyoti and Behl, Harkirat and Bubeck, S{\'e}bastien and Eldan, Ronen and Gunasekar, Suriya and Harrison, Michael and Hewett, Russell J and Javaheripi, Mojan and Kauffmann, Piero and others},
  journal={arXiv preprint arXiv:2412.08905},
  year={2024}
}

@article{shao2024deepseekmath,
  title={Deepseekmath: Pushing the limits of mathematical reasoning in open language models},
  author={Shao, Zhihong and Wang, Peiyi and Zhu, Qihao and Xu, Runxin and Song, Junxiao and Bi, Xiao and Zhang, Haowei and Zhang, Mingchuan and Li, YK and Wu, Yang and others},
  journal={arXiv preprint arXiv:2402.03300},
  year={2024}
}

@article{openai2024o1,
  title={A survey of frontiers in llm reasoning: Inference scaling, learning to reason, and agentic systems},
  author={Ke, Zixuan and Jiao, Fangkai and Ming, Yifei and Nguyen, Xuan-Phi and Xu, Austin and Long, Do Xuan and Li, Minzhi and Qin, Chengwei and Wang, Peifeng and Savarese, Silvio and others},
  journal={arXiv preprint arXiv:2504.09037},
  year={2025}
}

@article{snell2025scaling,
  title={Scaling llm test-time compute optimally can be more effective than scaling model parameters},
  author={Snell, Charlie and Lee, Jaehoon and Xu, Kelvin and Kumar, Aviral},
  journal={arXiv preprint arXiv:2408.03314},
  year={2024},
  url={https://arxiv.org/abs/2408.03314}
}

@inproceedings{li2025fttt,
  title={Learning to reason from feedback at test-time},
  author={Li, Yanyang and Lyu, Michael R and Wang, Liwei},
  booktitle={Proceedings of the 63rd Annual Meeting of the Association for Computational Linguistics (Volume 1: Long Papers)},
  pages={5241--5253},
  year={2025}
}

@article{guo2025deepseekr1,
  title={Deepseek-r1: Incentivizing reasoning capability in llms via reinforcement learning},
  author={Guo, Daya and Yang, Dejian and Zhang, Haowei and Song, Junxiao and Wang, Peiyi and Zhu, Qihao and Xu, Runxin and Zhang, Ruoyu and Ma, Shirong and Bi, Xiao and others},
  journal={arXiv preprint arXiv:2501.12948},
  year={2025}
}

@inproceedings{besta2024got,
  title={Graph of thoughts: Solving elaborate problems with large language models},
  author={Besta, Maciej and Blach, Nils and Kubicek, Ales and Gerstenberger, Robert and Podstawski, Michal and Gianinazzi, Lukas and Gajda, Joanna and Lehmann, Tomasz and Niewiadomski, Hubert and Nyczyk, Piotr and others},
  booktitle={Proceedings of the AAAI conference on artificial intelligence},
  volume={38},
  pages={17682--17690},
  year={2024}
}

@inproceedings{gou2024tora,
  title={Tora: A tool-integrated reasoning agent for mathematical problem solving},
  author={Gou, Zhibin and Shao, Zhihong and Gong, Yeyun and Shen, Yelong and Yang, Yujiu and Huang, Minlie and Duan, Nan and Chen, Weizhu},
  booktitle={International Conference on Learning Representations},
  year={2024},
  url={https://arxiv.org/abs/2309.17452}
}

@article{koh2024treesearch,
  title={Tree search for language model agents},
  author={Koh, Jing Yu and McAleer, Stephen and Fried, Daniel and Salakhutdinov, Ruslan},
  journal={arXiv preprint arXiv:2407.01476},
  year={2024}
}

@article{xie2024mcts,
  title={Monte carlo tree search boosts reasoning via iterative preference learning},
  author={Xie, Yuxi and Goyal, Anirudh and Zheng, Wenyue and Kan, Min-Yen and Lillicrap, Timothy P and Kawaguchi, Kenji and Shieh, Michael},
  journal={arXiv preprint arXiv:2405.00451},
  year={2024}
}

@article{zhao2024marco,
  title={Marco-o1: Towards open reasoning models for open-ended solutions},
  author={Zhao, Yu and Yin, Huifeng and Zeng, Bo and Wang, Hao and Shi, Tianqi and Lyu, Chenyang and Wang, Longyue and Luo, Weihua and Zhang, Kaifu},
  journal={arXiv preprint arXiv:2411.14405},
  year={2024}
}

@article{wang2024selfcorrection,
  title={A theoretical understanding of self-correction through in-context alignment},
  author={Wang, Yifei and Wu, Yuyang and Wei, Zeming and Jegelka, Stefanie and Wang, Yisen},
  journal={Advances in Neural Information Processing Systems},
  volume={37},
  pages={89869--89912},
  year={2024}
}

@inproceedings{muennighoff2025s1,
  title={s1: Simple test-time scaling},
  author={Muennighoff, Niklas and Yang, Zitong and Shi, Weijia and Li, Xiang Lisa and Fei-Fei, Li and Hajishirzi, Hannaneh and Zettlemoyer, Luke and Liang, Percy and Cand{\`e}s, Emmanuel and Hashimoto, Tatsunori B},
  booktitle={Proceedings of the 2025 Conference on Empirical Methods in Natural Language Processing},
  pages={20286--20332},
  year={2025}
}

@article{kimiteam2025k15,
  title={Kimi k1. 5: Scaling reinforcement learning with llms},
  author={Team, Kimi and Du, Angang and Gao, Bofei and Xing, Bowei and Jiang, Changjiu and Chen, Cheng and Li, Cheng and Xiao, Chenjun and Du, Chenzhuang and Liao, Chonghua and others},
  journal={arXiv preprint arXiv:2501.12599},
  year={2025}
}

@article{liu2024rmbench,
  title={Rm-bench: Benchmarking reward models of language models with subtlety and style},
  author={Liu, Yantao and Yao, Zijun and Min, Rui and Cao, Yixin and Hou, Lei and Li, Juanzi},
  journal={arXiv preprint arXiv:2410.16184},
  year={2024}
}

@article{luo2024automatedprocess,
  title={Improve mathematical reasoning in language models by automated process supervision},
  author={Luo, Liangchen and Liu, Yinxiao and Liu, Rosanne and Phatale, Samrat and Guo, Meiqi and Lara, Harsh and Li, Yunxuan and Shu, Lei and Zhu, Yun and Meng, Lei and others},
  journal={arXiv preprint arXiv:2406.06592},
  year={2024},
  url={https://arxiv.org/abs/2406.06592}
}

@inproceedings{zheng2024processbench,
  title={Processbench: Identifying process errors in mathematical reasoning},
  author={Zheng, Chujie and Zhang, Zhenru and Zhang, Beichen and Lin, Runji and Lu, Keming and Yu, Bowen and Liu, Dayiheng and Zhou, Jingren and Lin, Junyang},
  booktitle={Proceedings of the 63rd Annual Meeting of the Association for Computational Linguistics (Volume 1: Long Papers)},
  pages={1009--1024},
  year={2025}
}

@article{zhang2024gsm1k,
  title={A careful examination of large language model performance on grade school arithmetic},
  author={Zhang, Hugh and Da, Jeff and Lee, Dean and Robinson, Vaughn and Wu, Catherine and Song, Will and Zhao, Tiffany and Raja, Pranav and Slack, Dylan and Lyu, Qin and others},
  journal={arXiv preprint arXiv:2405.00332},
  year={2024},
  url={https://arxiv.org/abs/2405.00332}
}

@article{mirzadeh2024gsmsymbolic,
  title={Gsm-symbolic: Understanding the limitations of mathematical reasoning in large language models},
  author={Mirzadeh, Iman and Alizadeh, Keivan and Shahrokhi, Hooman and Tuzel, Oncel and Bengio, Samy and Farajtabar, Mehrdad},
  journal={arXiv preprint arXiv:2410.05229},
  year={2024}
}

@inproceedings{he2024olympiadbench,
  title={Olympiadbench: A challenging benchmark for promoting agi with olympiad-level bilingual multimodal scientific problems},
  author={He, Chaoqun and Luo, Renjie and Bai, Yuzhuo and Hu, Shengding and Thai, Zhen and Shen, Junhao and Hu, Jinyi and Han, Xu and Huang, Yujie and Zhang, Yuxiang and others},
  booktitle={Proceedings of the 62nd Annual Meeting of the Association for Computational Linguistics (Volume 1: Long Papers)},
  pages={3828--3850},
  year={2024}
}

@article{gao2024omnimath,
  title={Omni-math: A universal olympiad level mathematic benchmark for large language models},
  author={Gao, Bofei and Song, Feifan and Yang, Zhe and Cai, Zefan and Miao, Yibo and Dong, Qingxiu and Li, Lei and Ma, Chenghao and Chen, Liang and Xu, Runxin and others},
  journal={arXiv preprint arXiv:2410.07985},
  year={2024}
}

@article{glazer2024frontiermath,
  title={Frontiermath: A benchmark for evaluating advanced mathematical reasoning in ai},
  author={Glazer, Elliot and Erdil, Ege and Besiroglu, Tamay and Chicharro, Diego and Chen, Evan and Gunning, Alex and Olsson, Caroline Falkman and Denain, Jean-Stanislas and Ho, Anson and Santos, Emily de Oliveira and others},
  journal={arXiv preprint arXiv:2411.04872},
  year={2024}
}

@article{deepseek2024v3,
  title={Deepseek-v3 technical report},
  author={Liu, Aixin and Feng, Bei and Xue, Bing and Wang, Bingxuan and Wu, Bochao and Lu, Chengda and Zhao, Chenggang and Deng, Chengqi and Zhang, Chenyu and Ruan, Chong and others},
  journal={arXiv preprint arXiv:2412.19437},
  year={2024}
}

@misc{qwen2024qwq,
  title={Qwq: Reflect deeply on the boundaries of the unknown, November 2024},
  author={Qwen, Team},
  url={https://qwenlm.github.io/blog/qwq-32b-preview/},
  year={2024}
}

@article{liu2025computeoptimal,
  title={Can 1b llm surpass 405b llm? rethinking compute-optimal test-time scaling},
  author={Liu, Runze and Gao, Junqi and Zhao, Jian and Zhang, Kaiyan and Li, Xiu and Qi, Biqing and Ouyang, Wanli and Zhou, Bowen},
  journal={arXiv preprint arXiv:2502.06703},
  year={2025},
  url={https://arxiv.org/abs/2502.06703}
}

@article{ye2025limo,
  title={Limo: Less is more for reasoning},
  author={Ye, Yixin and Huang, Zhen and Xiao, Yang and Chern, Ethan and Xia, Shijie and Liu, Pengfei},
  journal={arXiv preprint arXiv:2502.03387},
  year={2025}
}

@article{openthoughts2025data,
  title={Openthoughts: Data recipes for reasoning models},
  author={Guha, Etash and Marten, Ryan and Keh, Sedrick and Raoof, Negin and Smyrnis, Georgios and Bansal, Hritik and Nezhurina, Marianna and Mercat, Jean and Vu, Trung and Sprague, Zayne and others},
  journal={arXiv preprint arXiv:2506.04178},
  year={2025},
  url={https://arxiv.org/abs/2506.04178}
}

@article{qwen2025qwen3,
  title={Qwen3 technical report},
  author={Yang, An and Li, Anfeng and Yang, Baosong and Zhang, Beichen and Hui, Binyuan and Zheng, Bo and Yu, Bowen and Gao, Chang and Huang, Chengen and Lv, Chenxu and others},
  journal={arXiv preprint arXiv:2505.09388},
  year={2025}
}

@article{song2025rewardenough,
  title={Reward is enough: Llms are in-context reinforcement learners},
  author={Song, Kefan and Moeini, Amir and Wang, Peng and Gong, Lei and Chandra, Rohan and Zhang, Shangtong and Qi, Yanjun},
  journal={arXiv preprint arXiv:2506.06303},
  year={2025}
}

@inproceedings{sinha2026diminishingreturns,
  title={The illusion of diminishing returns: Measuring long horizon execution in llms},
  author={Sinha, Akshit and Arun, Arvindh and Goel, Shashwat and Staab, Steffen and Geiping, Jonas},
  booktitle={International Conference on Learning Representations},
  year={2026},
  url={https://arxiv.org/abs/2509.09677}
}

@article{ahmed2026retrievalthought,
  title={Retrieval-of-Thought: Efficient Reasoning via Reusing Thoughts},
  author={Ahmed, Ammar and Khan, Azal Ahmad and Ahmad, Ayaan and Di, Sheng and Liu, Zirui and Anwar, Ali},
  journal={arXiv preprint arXiv:2509.21743},
  year={2025},
  url={https://arxiv.org/abs/2509.21743}
}

@misc{clark2018thinksolvedquestionanswering,
      title={Think you have Solved Question Answering? Try ARC, the AI2 Reasoning Challenge}, 
      author={Peter Clark and Isaac Cowhey and Oren Etzioni and Tushar Khot and Ashish Sabharwal and Carissa Schoenick and Oyvind Tafjord},
      year={2018},
      eprint={1803.05457},
      archivePrefix={arXiv},
      primaryClass={cs.AI},
      url={https://arxiv.org/abs/1803.05457}, 
}


\appendix

\section{Experimental Protocol, Data Scope, and Reproducibility}
\label{app:protocol}

The main experiments use a generate-then-rerank protocol. Aside from the candidate-construction layer, comparisons keep benchmark split, retained cap, parser, scoring endpoint, and downstream selector fixed. Numeric tasks use final numeric exact match, while MCQ tasks are hard-locked to \Choice{}. Table~\ref{tab:main-results} in the main text reports the main results, and Appendix Tables~\ref{tab:decomposition} and~\ref{tab:table4} report auxiliary decomposition and interaction evidence. This appendix consolidates the experimental protocol, data scope, statistical conventions, and reproducibility interfaces, ensuring that every reported number traces back to a run manifest, prediction record, or aggregate diagnostic summary.

The model stack is as follows. The main candidate generator is Qwen3-1.7B. The fixed selector in the main experiments is a Qwen3-1.7B model fine-tuned on GSM8K training-split candidate pairs with a binary yes/no objective (checkpoint id \texttt{qwen3\_17b\_gsm8k512\_verifier\_yesno\_v1}). The verifier was not trained on MMLU-Pro, GPQA, or any evaluation split used in this paper. \CACC{}+\SPP{} rows use a Qwen3-8B proposer / prompt-mixture variant for proposer-side strengthening. The FTTT reference uses its official test-time feedback/training protocol and is therefore reported as a high-compute external reference. Table~\ref{tab:app-run-ledger} summarizes split, answer mode, and paper use for each evidence unit.

\begin{table}[!ht]
\centering
\footnotesize
\caption{Evidence units, sizes, and core metrics. The table lists sample size, answer mode, oracle, final, and V|O for the main evidence units in the main text.}
\label{tab:app-run-ledger}
\setlength{\tabcolsep}{2.4pt}
\begin{tabular}{@{}l r l c c c l@{}}
\toprule
Evidence unit & $N$ & Mode & Oracle & Final & V|O & Use in paper \\
\midrule
GSM8K Base & 1319 & numeric & 644/1319 & 379/1319 & 0.5885 & Numeric baseline \\
GSM8K \CACC{} & 1319 & numeric & 843/1319 & 479/1319 & 0.5682 & Numeric main result \\
CompMath Base & 3199 & numeric & 1027/3199 & 486/3199 & 0.4732 & Numeric baseline \\
CompMath \CACC{} & 3199 & numeric & 1419/3199 & 825/3199 & 0.5814 & Numeric main result \\
MMLU no-\AMC{} numeric & 12032 & numeric & 444/12032 & 115/12032 & 0.2590 & Archived collapse provenance \\
MMLU MCQ reparse & 12032 & \Choice{} & 4900/12032 & 2842/12032 & 0.5800 & Readout recovery \\
MMLU \CACC{} & 12032 & \Choice{} & 5736/12032 & 2740/12032 & 0.4777 & MCQ Gap exposure \\
MMLU \CACC{}+\SPP{} & 12032 & \Choice{} & 6650/12032 & 3532/12032 & 0.5311 & Proposer-side narrowing \\
GPQA Base & 198 & \Choice{} & 38/198 & 11/198 & 0.2897 & MCQ Baseline \\
GPQA \CACC{} & 198 & \Choice{} & 90/198 & 38/198 & 0.4222 & MCQ Compatibility result \\
GPQA \CACC{}+\SPP{} & 198 & \Choice{} & 97/198 & 51/198 & 0.5258 & Proposer-side narrowing \\
Power Sampling x \CACC{} & 1319 & numeric & 0.7195 & 0.3381 & 0.4700 & Diagnostic widening case \\
\bottomrule
\end{tabular}
\end{table}

Data and statistical records are saved together with the run manifest. Each manifest contains at least the dataset/split id, sample filtering rules, candidate source, retained cap, parser endpoint, scoring script, selector checkpoint, random seed, prediction JSONL path, and figure/table generation script, so that every number in the main text traces back to a prediction record or aggregate diagnostic summary. Paired closure and rescue rows can recompute McNemar tests and bootstrap confidence intervals, while Power Sampling x \CACC{} is used as a summary-only diagnostic.

Table~\ref{tab:app-compute} summarizes the inference operations in the audited configuration. \CACC{} adds one batched generation of four completion sequences per example and rescoring of the modified candidate pool. The \SPP{} rows use a Qwen3-8B proposer.

\begin{table}[!ht]
\centering
\footnotesize
\caption{Inference operations in the audited configuration.}
\label{tab:app-compute}
\begin{tabularx}{\linewidth}{@{}l X@{}}
\toprule Component & Work and accounting boundary\\\midrule
Upstream generation & Audited contract: 16 sequences, at most 128 new tokens each; pool construction precedes retention.\\
Completion construction & One additional batch of 4 sequences, at most 160 new tokens each, per example.\\
Routing / normalization & Deterministic diagnostics, parsing, sanitization, and deduplication.\\
Fixed selector & Two yes/no continuation scores per retained candidate; batching determines forward-pass count.\\
Stronger proposer & Qwen3-8B replaces or augments the 1.7B source in the composition setting; accounted separately.\\
\bottomrule\end{tabularx}
\end{table}

Specifically, GSM8K clean / full, \texttt{competition\_math:numeric}, MMLU-Pro test, GPQA Diamond, MMLU-Pro confirm-512, and GPQA-128 each record split source, sample size, filtering rules, parser endpoint, and run id; smoke evidence and full/confirm evidence are tagged separately to maintain traceability across evidence scales. Paired closure uses McNemar tests with bootstrap confidence intervals. Code, configurations, and reproduction instructions are available at \url{https://github.com/drongzzz0/CACC-nips}. This work studies the candidate-interface problem in small-model reasoning enhancement and does not involve personal data decisions or high-risk automated deployment; its main application boundary is low-cost, local, or intranet small-model reasoning systems.

\begin{table}[!ht]
\centering
\footnotesize
\caption{GSM8K candidate-budget diagnostics: oracle, V|O, and final do not move synchronously when sample count or retained cap changes.}
\label{tab:app-budget-diagnostics}
\setlength{\tabcolsep}{3pt}
\begin{tabularx}{\linewidth}{@{}l l c c c c Y@{}}
\toprule
Diagnostic & Setting & Oracle & V|O & Final & $\Delta F$ & Note \\
\midrule
Sample sweep & cap 8, samples 4 & 0.6422 & 0.5844 & 0.3753 & ref. & Fewer samples but high conversion \\
Sample sweep & cap 8, samples 32 & 0.6429 & 0.4517 & 0.2904 & -8.49pp & Similar oracle, lower conversion \\
Retained-cap sweep & cap 8 & 0.6391 & 0.5682 & 0.3632 & ref. & CACC canonical retained cap \\
Retained-cap sweep & cap 16 & 0.8249 & 0.3998 & 0.3298 & -3.34pp & Larger retained pool raises oracle but hurts conversion \\
\bottomrule
\end{tabularx}
\end{table}

\begin{table}[!ht]
\centering
\footnotesize
\caption{GSM8K same-pool verifier sensitivity (final-only).}
\label{tab:app-verifier-sensitivity}
\setlength{\tabcolsep}{5pt}
\begin{tabular}{@{}l c c l@{}}
\toprule
Setting & Final & $\Delta F$ & Note \\
\midrule
Canonical verifier & 0.3632 & ref. & CACC canonical selector \\
verifier128 & 0.3465 & -1.67pp & Same-pool selector sensitivity check \\
verifier512 & 0.3594 & -0.38pp & close to Canonical verifier final \\
\bottomrule
\end{tabular}
\end{table}

\section{Coverage Decomposition and Parser Provenance}
\label{app:mmlu-provenance}

The auxiliary decomposition in Table~\ref{tab:decomposition} (main text) reports paired deltas across audit rows. All deltas are subtractions over the same set of frozen rows, with no extra experiments; positive final gains can come from oracle gain or V|O gain. The MMLU-Pro \CACC{} row is the key mechanism point: even after oracle rises, answer-mode alignment and proposer-side strengthening are still required to convert it stably into final gain. Power Sampling x \CACC{} is used as a coverage-pressure diagnostic, illustrating that raw coverage, V|O, and final must be tracked jointly.

\paragraph{MMLU-Pro parser, scoring, and reparse provenance.}

This subsection documents the parser and scoring provenance for the MMLU-Pro compatibility evidence. The \texttt{CCR-only} artifact corresponds to a legacy-mode candidate pool whose metadata remains numeric. Under the archived numeric exact-match scoring path, this legacy/no-\AMC{} candidate-interface setting yields oracle \texttt{444/12032 = 0.0369} and verifier \texttt{115/12032 = 0.0096}. The benchmark-locked MCQ reparse reads the same legacy/no-\AMC{} candidate-interface as oracle \texttt{4900/12032} and verifier \texttt{2842/12032}, supporting the AMC/readout chain in the main text. Table~\ref{tab:app-mmlu-provenance} summarizes this provenance.

\begin{table}[!ht]
\centering
\footnotesize
\caption{MMLU-Pro parser and scoring provenance: this comparison is a scoring/readout-path ablation, not an ablation between two independent candidate pools.}
\label{tab:app-mmlu-provenance}
\setlength{\tabcolsep}{2.2pt}
\begin{tabularx}{\linewidth}{@{}l l r c c c X@{}}
\toprule
View & Parser / scoring path & $N$ & Final & Oracle & V|O & Use \\
\midrule
legacy/no-\AMC{} & archived numeric exact match & 12032 & 115/12032 & 444/12032 & 0.2590 & Numeric scoring near-collapse \\
legacy/no-\AMC{} & benchmark-locked MCQ reparse & 12032 & 2842/12032 & 4900/12032 & 0.5800 & Readout recovery \\
CCR+\AMC{} & benchmark-locked MCQ reparse & 12032 & 2731/12032 & 5745/12032 & 0.4754 & Micro-ablation reference \\
\CACC{} & benchmark-locked MCQ reparse & 12032 & 2740/12032 & 5736/12032 & 0.4777 & Gap exposure \\
\CACC{}+\SPP{} & benchmark-locked MCQ reparse & 12032 & 3532/12032 & 6650/12032 & 0.5311 & Gap narrowing \\
\bottomrule
\end{tabularx}
\end{table}

The MCQ scoreboard in this paper uses benchmark-locked choice-letter reparsing. MMLU-Pro and GPQA are both hard-locked to the \Choice{} answer mode. This comparison therefore tests whether the legacy/no-\AMC{} candidate-interface becomes usable under the correct MCQ readout path, and is presented in the main text as a scoring/readout-path ablation.

\section{Recovery, Composition, and Interaction Evidence}

The ablation/recovery/interaction matrix is reported in Table~\ref{tab:table4} (main text); this appendix expands the supporting details.

\begin{figure}[!ht]
\centering
\includegraphics[width=0.96\linewidth]{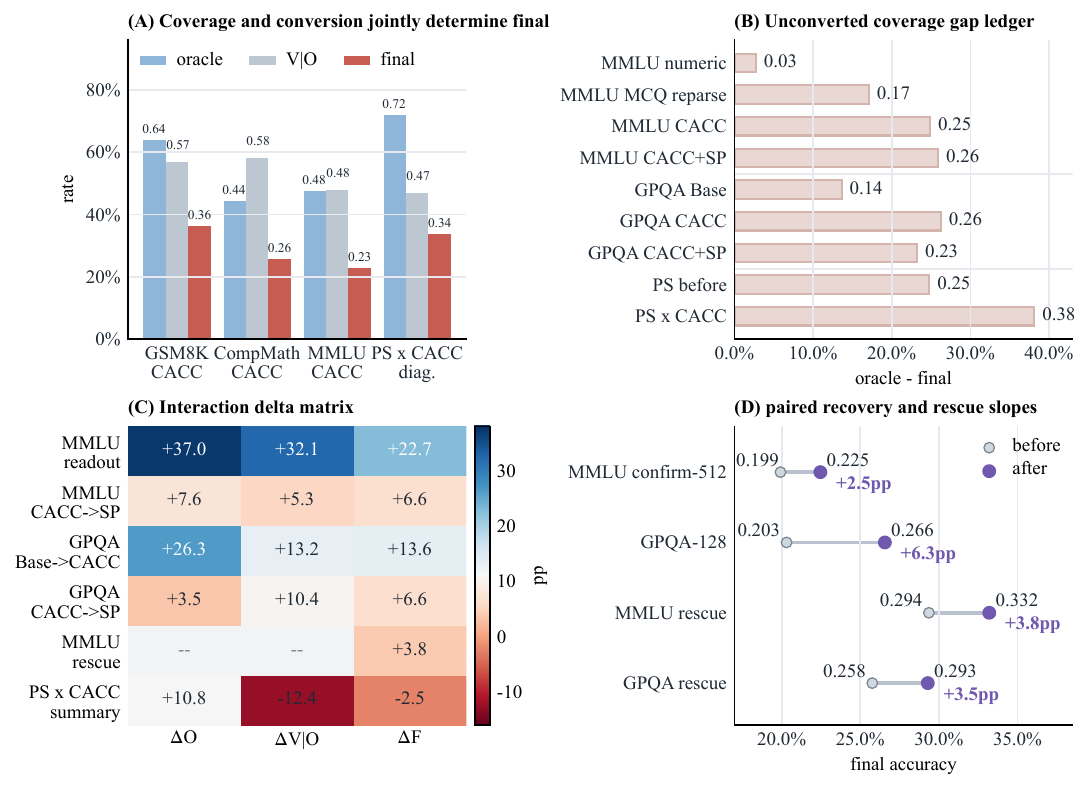}
\caption{Mechanistic analysis of the coverage--conversion gap, compatibility, policy tradeoffs, and recovery closure. Panel A: oracle, V|O, and final decomposition. Panel B: unconverted gap between oracle and final. Panel C: interaction delta matrices for $\Delta$oracle, $\Delta$V|O, and $\Delta$final. Panel D: before$\rightarrow$after slopes. Paired statistics are in Tables~\ref{tab:table4} and~\ref{tab:app-paired-details}.}
\label{fig:mechanism}
\end{figure}

\begin{figure}[!ht]
\centering
\includegraphics[width=0.72\linewidth]{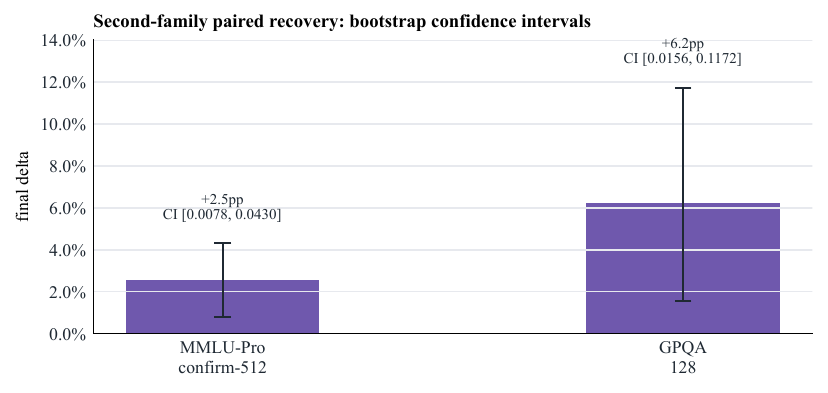}
\caption{Bootstrap confidence intervals for second-family paired recovery. Both closure intervals lie above zero, supporting recoverable compatibility.}
\label{fig:app-second-family-ci}
\end{figure}

The main Second-family recovery evidence is in the recovery rows of Table~\ref{tab:table4} and Panel D of Figure~\ref{fig:mechanism}; Figure~\ref{fig:app-second-family-ci} additionally visualizes the paired bootstrap intervals. Reproducing this analysis requires retaining paired example ids, before/after prediction JSONL, McNemar contingency counts, and bootstrap seed, so that the $p$-values and confidence intervals reported in the main text can be recomputed. The result supports recoverable second-family compatibility.

\begin{table}[!ht]
\centering
\footnotesize
\caption{Paired statistics and diagnostic details for the mechanism rows in Table~\ref{tab:table4}.}
\label{tab:app-paired-details}
\setlength{\tabcolsep}{2.0pt}
\begin{tabularx}{\linewidth}{@{}Y l c c r c Y Y@{}}
\toprule
Evidence unit & Split / scope & Before & After & $\Delta F$ & w/l & Statistical detail & Artifact \\
\midrule
Second-family recovery & MMLU confirm-512 & 102/512 (0.1992) & 115/512 (0.2246) & +2.54pp & 17/4 & McNemar $p=0.007197$; CI [0.0078, 0.0430] & Paired JSONL \\
Second-family recovery & GPQA-128 & 26/128 (0.2031) & 34/128 (0.2656) & +6.25pp & 9/1 & McNemar $p=0.02148$; CI [0.0156, 0.1172] & Paired JSONL \\
Selection rescue & MMLU-Pro & 0.2936 & 0.3320 & +3.84pp & 613/150 & McNemar $p=3.43{\times}10^{-67}$ & Paired pred. \\
Selection rescue & GPQA & 0.2576 & 0.2929 & +3.53pp & 10/3 & McNemar $p=0.0923$ & Paired pred. \\
Coverage pressure & GSM8K full & 0.3632 & 0.3381 & -2.50pp & summary & Oracle +10.84pp; V|O -12.43pp & Summary only \\
Second-family smoke & GSM8K-128 & 27/128 (0.2109) & 29/128 (0.2266) & +1.56pp & 7/1 & Smoke row & Paired rows available \\
Second-family smoke & MMLU-128 & 24/128 (0.1875) & 27/128 (0.2109) & +2.34pp & 8/4 & Smoke row & Paired rows \\
\bottomrule
\end{tabularx}
\end{table}

\paragraph{Layered composition and interaction data.}
\label{app:layer-aware}

This subsection documents the evidence scope of the layer-aware composition. Self-Refine corresponds to the trajectory-improvement layer, self-consistency to the candidate-source layer, ToT/search to the reasoning-space-expansion layer, FTTT to the feedback/test-time-training layer, and verifier/reranker to the selection layer. \CACC{} corresponds to the candidate-interface layer. Table~\ref{tab:table4} reports the interaction evidence in the main text, and Table~\ref{tab:app-artifacts} further distinguishes frozen statistical results from summary-only diagnostics.

\begin{table}[!ht]
\centering
\footnotesize
\caption{Artifact availability matrix: artifact completeness and paper use of each evidence unit.}
\label{tab:app-artifacts}
\setlength{\tabcolsep}{3.2pt}
\begin{tabularx}{\linewidth}{@{}l c c c Y@{}}
\toprule
Evidence unit & Candidate pool & Prediction records & Paired statistics & Manuscript use \\
\midrule
Main Base / \CACC{} rows & available & available & row-level deltas & Frozen result rows \\
MMLU no-\AMC{} numeric view & archived & available & provenance audit & Scoring/readout-path provenance \\
MMLU MCQ reparse & archived scoreboard & available & provenance audit & MCQ Readout recovery \\
Second-family closure & available & available & McNemar + bootstrap & Frozen paired recovery \\
Conversion-aware rescue & available & available & McNemar where available & Gap narrowing evidence \\
Power Sampling x \CACC{} & summary + pool & summary metrics & aggregate diagnostic & Diagnostic summary-only \\
\bottomrule
\end{tabularx}
\end{table}

\textbf{FTTT high-compute external reference.}
The archived FTTT GSM8K full log reports $1152/1319 = 0.8734$ with an average of $6.67$ test-time trials. This reading corresponds to the official test-time feedback/training protocol; its sample budget, feedback loop, and model-state updates fall in a high-compute feedback/test-time-training regime. We use it as an external high-compute reference to illustrate the system-position difference between the candidate-interface layer and the feedback/test-time-training layer.

\begin{table}[!ht]
\centering
\footnotesize
\caption{External baseline provenance and artifact status.}
\label{tab:app-external-provenance}
\setlength{\tabcolsep}{2.8pt}
\begin{tabularx}{\linewidth}{@{}l l c c c X@{}}
\toprule
Method & Task & Correct / $N$ & Final & Extra cost & Artifact status \\
\midrule
Self-Refine & GSM8K full & 643/1319 & 0.4875 & 5.0 calls/example & Full row available \\
Self-Refine & CompMath numeric & 789/3199 & 0.2466 & 17.0 calls/example & Full row available \\
Self-Refine & MMLU-Pro & 2838/12032 & 0.2359 & 17.0 calls/example & Full row available \\
Self-Refine & GPQA Diamond & 24/198 & 0.1212 & 17.0 calls/example & Full row available \\
Self-Refine & GSM8K-128 & 51/128 & 0.3984 & eval-128 & Small reference row \\
ToT-lite & GSM8K-128 & 37/128 & 0.2891 & eval-128 & Small search baseline row \\
CCQA & GSM8K-128 & 21/128 & 0.1641 & eval-128 & Small query-generation row \\
FTTT & GSM8K full & 1152/1319 & 0.8734 & 6.67 trials/example & Final-only high-compute reference \\
PairRM & CompMath numeric & 257/3199 & 0.0803 & best-of-16 & Primary report \\
PairRM & CompMath numeric & 244/3199 & 0.0763 & best-of-16 & Secondary report; provenance discrepancy noted \\
PairRM & MMLU-Pro & 793/12032 & 0.0659 & best-of-8 & Primary report \\
\bottomrule
\end{tabularx}
\end{table}

\FloatBarrier
\section{Offline Rulebook Components}
\label{app:algorithm}

The policy-parameterized pseudocode of \CACC{} is given as Algorithm~\ref{alg:cacc} in Section~\ref{sec:method}. The offline rulebook components instantiated in our implementation are summarized below. The replacement policy is treated as a swappable component, while the method composition of \CACC{} remains \CCR{}+\AMC{}+\SCH{}. This is consistent with the offline--online flow in Figure~\ref{fig:method}: offline calibration produces a frozen rulebook, and online inference only executes gold-free routing and candidate construction.

\begin{table}[!ht]
\centering
\footnotesize
\caption{Offline rulebook components used by \CACC{}.}
\label{tab:app-rulebook}
\setlength{\tabcolsep}{3pt}
\begin{tabularx}{\linewidth}{@{}l X X X@{}}
\toprule
Component & Diagnostic trigger & Online action & Frozen output \\
\midrule
\CCR{} & partial derivation, missing final answer, fragmented numeric form & repair final-answer span and recover parseable candidate text & candidate text plus parser status \\
\AMC{} & Answer-mode mismatch, missing choice letter, Wrapper drift & map candidate to benchmark-locked mode and protect choice-letter wrapper & mode-aligned candidate \\
\SCH{} & duplicated scaffold, template residue, selector-attractive noise & remove residue, deduplicate, and keep retained cap stable & hygienic retained pool \\
Routing policy & diagnostic vector $\phi$ and retained cap $K$ & select rule sequence without gold access & frozen $\pi(\phi,K)$ \\
\bottomrule
\end{tabularx}
\end{table}

\FloatBarrier
\section{Interface Error Taxonomy and Rule Interfaces}

Table~\ref{tab:app-error-taxonomy} lists the candidate-interface error classes used in our qualitative analysis.

\begin{table}[!ht]
\centering
\footnotesize
\caption{Candidate-interface error taxonomy used in the qualitative analysis.}
\label{tab:app-error-taxonomy}
\begin{tabularx}{\linewidth}{l X l}
\toprule
Error class & Diagnostic signal & Expected effect \\
\midrule
Oracle miss & No candidate parses to gold & Low oracle \\
Fragment / partial & Reasoning stops before final answer & Low oracle / Low V|O \\
Answer-mode mismatch & Numeric answer in MCQ or missing choice letter & Low MCQ final \\
Invalid final & Malformed final channel & Parser failure \\
Scaffold residue & Instruction or template leakage & Selector confusion \\
Wrapper drift & Second-family answer wrapper differs & Transfer loss \\
Oracle-hit verifier miss & Correct candidate present but not selected & Low V|O \\
\bottomrule
\end{tabularx}
\end{table}

This taxonomy makes explicit the difference between raw candidate presence and selectable coverage. Each error class corresponds to one path of candidate-interface failure, showing how a candidate can affect the final decision through Parser failure, Answer-mode mismatch, or Selector confusion.

\begin{table}[!ht]
\centering
\footnotesize
\caption{Representative audit examples. These examples are used for post-hoc qualitative analysis only, not for offline rule calibration.}
\label{tab:app-audit-examples}
\setlength{\tabcolsep}{2.5pt}
\begin{tabularx}{\linewidth}{@{}l l l X@{}}
\toprule
Example id & Dataset & Gold & Audited failure / recovery pattern \\
\midrule
\texttt{gsm8k-test-00595} & GSM8K & 13 & Oracle miss; \CACC{} verifier selects 15 although gold is 13 \\
\texttt{competition-math-numeric-test-02878} & CompMath & 13 & raw oracle hit exists, but verifier overrules first candidate and selects 11 \\
\texttt{mmlu-pro-test-08795} & MMLU-Pro & A & no-\AMC{} numeric view misses a choice-form solution; MCQ-aware \CACC{}+\SPP{} later recovers final answer A \\
\texttt{gsm8k-test-00462} & GSM8K & 71 & \CACC{} repairs Fragment / partial candidate into a boxed final answer \\
\texttt{gpqa-diamond-train-recNFJjE5PPTqVJGv} & GPQA & B & selected wrong candidate contains scaffold / instruction residue, illustrating hygiene pressure \\
\texttt{mmlu-pro-test-08026} & MMLU-Pro & D & stronger proposer adds/selects D after CCR/no-\AMC{} misses the answer \\
\texttt{mmlu-pro-test-00942} & MMLU-Pro & F & second-family CACC selects long option text; hygiene recovers bare choice letter F \\
\bottomrule
\end{tabularx}
\end{table}

\paragraph{Prompts, templates, parsers, and sanitizers.}

This subsection records the reproducibility interfaces of the \CCR{} prompt template, \AMC{} answer wrapper, MCQ choice-letter lock, numeric parser normalization, sanitizer rules, final answer extraction, and benchmark-specific prompt notes. All prompt and parser entries are bound to the run manifest, simplifying reproduction of the experimental candidate construction and scoring path.

\begin{table}[!ht]
\centering
\footnotesize
\caption{Prompt, parser, and sanitizer interfaces bound to run manifests.}
\label{tab:app-parser-rules}
\setlength{\tabcolsep}{3pt}
\begin{tabularx}{\linewidth}{@{}l X X@{}}
\toprule
Interface item & Constraint & Reproducibility role \\
\midrule
\CCR{} prompt template & no gold access; repair only observable fragments & reproduces Fragment / partial candidate repair \\
\AMC{} answer wrapper & numeric tasks keep exact numeric form; MCQ tasks hard-lock to \Choice{} & fixes answer-mode alignment and readout path \\
Numeric parser normalization & normalize whitespace, delimiters, final numeric span, and wrapper variants & keeps open-ended scoring stable across candidate sources \\
MCQ parser configuration & preserve benchmark option set and final choice-letter endpoint & supports MMLU-Pro / GPQA reparse provenance \\
\SCH{} sanitizer rules & remove Scaffold residue, duplicated wrappers, and selector-attractive debris & stabilizes verifier input distribution \\
Manifest binding & every prompt/parser/sanitizer version is tied to run id, seed, split, and scorer & enables rerunning table and figure generation scripts \\
\bottomrule
\end{tabularx}
\end{table}

\noindent\textbf{Evidence scope and claim mapping.}
Table~\ref{tab:app-claim-map} maps the main claims (numeric final gain, MCQ answer-mode alignment, Second-family recovery, proposer-side strengthening, selection-side rescue, and coverage-pressure diagnostic) to the evidence blocks of Tables~\ref{tab:main-results}, \ref{tab:decomposition}, and~\ref{tab:table4}.

\begin{table}[H]
\centering
\footnotesize
\caption{Evidence scope and claim mapping.}
\label{tab:app-claim-map}
\begin{tabularx}{\linewidth}{Y c Y}
\toprule
Claim & Evidence unit & Paper use \\
\midrule
\CACC{} improves numeric open-ended reasoning & GSM8K + competition\_math & Direct final-gain evidence \\
MCQ benefits from \AMC{} / scoring-path alignment & MMLU-Pro readout chain & Answer-mode alignment evidence \\
Second-family compatibility is recoverable & MMLU confirm-512 / GPQA-128 & Paired recovery evidence \\
\CACC{}+\SPP{} narrows MCQ gap & MMLU-Pro / GPQA & Proposer-side composition evidence \\
Conversion-aware rescue improves final selection & MMLU-Pro / GPQA & Selection-side composition evidence \\
Power Sampling x \CACC{} diagnoses Coverage pressure & aggregate summary & Coverage--conversion diagnostic \\
\bottomrule
\end{tabularx}
\end{table}

This mapping separates frozen results, paired recovery, and diagnostic summaries by their paper roles, ensuring a one-to-one correspondence between main-text claims and auditable artifacts.

\FloatBarrier
\section{Notation Summary and Per-Benchmark Setup}
\label{app:notation}

Table~\ref{tab:app-notation} consolidates the symbols used across the paper, and Table~\ref{tab:app-benchmarks} summarizes per-benchmark setup details (split, parser, scoring endpoint, sample budget) so that each row in the main results can be reproduced from the manuscript without prior context.

\begin{table}[H]
\centering
\footnotesize
\caption{Notation summary. Symbols are introduced in Sections~3 and~4 and used throughout the paper.}
\label{tab:app-notation}
\setlength{\tabcolsep}{4pt}
\begin{tabularx}{\linewidth}{@{}l X@{}}
\toprule
Symbol & Meaning \\
\midrule
$x$, $y_x$, $\mathcal D$ & input instance, ground-truth answer, evaluation dataset \\
$m\in\{\text{numeric},\Choice{}\}$ & benchmark-locked answer mode, fixed by registry $\mathcal M$ \\
$C_0(x)=\{c_1,\ldots,c_n\}$ & raw candidate pool of size $n$ produced by the generator \\
$T_m$, $R_K$, $C_K(x,m)$ & mode-aware candidate construction, retained-cap controller, retained pool of size $\le K$ \\
$\phi(x,c,m)$ & gold-free diagnostic feature vector (parseability, Answer-mode mismatch, residue, \ldots) \\
$\mathcal R$, $\pi$ & frozen rulebook ($\CCR{}/\AMC{}/\SCH{}$) and routing policy $\pi(\phi,K)\!\mapsto a$ \\
$S$, $\hat c$, $\hat y$ & fixed selector (verifier/reranker), selected candidate, parsed final answer \\
$\Afirst,\Aoracle,\Avo,\Afinal$ & first-answer accuracy, oracle coverage, conditional V|O, final accuracy \\
$\Aoracle^{\mathrm{sel}}$, $H_{\mathrm{sel}}$ & selectable oracle coverage and the selectable-correct hit event (Eq.~\ref{eq:hsel}) \\
\CACC{}, \CCR{}, \AMC{}, \SCH{} & method, completion-centric replacement, answer-mode compatibility, selector-compatible hygiene \\
\SPP{} & proposer-side strengthening (Qwen3-8B prompt-mixture proposer) used in composition rows \\
\bottomrule
\end{tabularx}
\end{table}

\begin{table}[H]
\centering
\footnotesize
\caption{Per-benchmark setup details. The retained cap is shared across compared methods on each benchmark.}
\label{tab:app-benchmarks}
\setlength{\tabcolsep}{1.5pt}
\begin{tabularx}{\linewidth}{@{}Y Y l l Y@{}}
\toprule
Benchmark & Split / $N$ & Mode & Retained cap & Parser / scoring endpoint \\
\midrule
GSM8K & clean test, $N{=}1319$ & numeric & 8 & numeric exact match on final boxed span \\
\texttt{competition\_\allowbreak math:numeric} & numeric subset, $N{=}3199$ & numeric & 8 & numeric exact match on normalized final span \\
MMLU-Pro & test, $N{=}12032$ & MCQ & 8 & benchmark-locked choice-letter reparse \\
GPQA Diamond & full, $N{=}198$ & MCQ & 8 & benchmark-locked choice-letter reparse \\
MMLU-Pro confirm-512 & confirm split, $N{=}512$ & MCQ & 8 & second-family paired closure run \\
GPQA-128 & subset, $N{=}128$ & MCQ & 8 & second-family paired closure run \\
GSM8K-128 / MMLU-128 & smoke subsets, $N{=}128$ & numeric/MCQ & 8 & Mistral-7B Second-family smoke runs \\
\bottomrule
\end{tabularx}
\end{table}

The two tables together with the run-ledger in Table~\ref{tab:app-run-ledger} make every row in Tables~\ref{tab:main-results}--\ref{tab:table4} traceable to its split, parser, scoring endpoint, sample budget, and answer mode.


\section{Controlled Evaluation Protocols and Detailed Results}
\label{app:controls}

\subsection{Experimental units and reporting conventions}
We identify each result by candidate pool, selector, readout, and evaluation set. Original-pool controls reuse the candidate contents underlying the historical results; fresh-generation controls create new pools and have separate baselines. Accuracies are measured per question. Three-seed verifier means are descriptive summaries, with per-seed paired intervals reported below. We do not treat repeated seeds on the same questions as independent question samples.

\subsection{Dataset-specific verifier supervision}
Verifiers use Qwen3-1.7B with response-only yes/no loss. MMLU-Pro supervision is restricted to the official validation set (70 questions, split into 56 train and 14 dev before format expansion). Evaluation on the 11,520-question held-out subset excludes the inspected confirm-512 subset. GPQA supervision comes from Extended after excluding Diamond overlaps using record identifiers and normalized question--option hashes; the cleaned set has 346 questions (278 train, 68 dev). Canonical and format-diverse option renderings are balanced 1:1 between yes and no within each format and split. Training uses three epochs, learning rate $2\times10^{-4}$, LoRA rank 8, alpha 16, dropout 0.05, batch size 1, gradient accumulation 4, and maximum length 768. Prompt/padding labels are masked; truncation is forbidden. Checkpoints are selected on grouped dev accuracy, then dev loss and binary accuracy, without evaluation labels. The fixed seeds are 14251, 20260724, and 314159.

Raw readout scores candidate text. Canonical readout maps a supported option to a common \texttt{Final answer: <LETTER>} representation and reuses its score across candidates supporting that option. It neither regenerates candidate reasoning nor adds a previously unsupported answer option. Within each target and seed, Base and \CACC{} use the same verifier. The GPQA canonical means are $16.16\%$ and $24.92\%$; MMLU held-out means are $26.40\%$ and $27.84\%$. Mean raw-readout gains are $9.93$\,pp and $0.37$\,pp, respectively, illustrating the effect of the readout contract.

\begin{table}[!ht]
\centering\footnotesize
\caption{Target-verifier canonical results on original pools. CIs are paired per-seed 95\% bootstrap intervals for $\Delta F$ (pp); $p$ is the per-contrast exact McNemar value, without a cross-row multiplicity adjustment.}
\label{tab:target-verifier-seeds}
\setlength{\tabcolsep}{4pt}
\begin{tabular}{@{}l r r r r r l l@{}}
\toprule
Evaluation & $N$ & Seed & Base & \CACC{} & $\Delta F$ & 95\% CI & $p$\\\midrule
GPQA & 198 & 14251 & 15.66 & 24.24 & +8.59 & [2.53,15.15] & .0137\\
GPQA & 198 & 20260724 & 16.67 & 26.26 & +9.60 & [3.54,15.66] & .00432\\
GPQA & 198 & 314159 & 16.16 & 24.24 & +8.08 & [1.52,14.65] & .0259\\
MMLU held-out & 11,520 & 14251 & 26.35 & 28.07 & +1.73 & [1.04,2.43] & $6.48\times10^{-7}$\\
MMLU held-out & 11,520 & 20260724 & 26.52 & 28.15 & +1.63 & [0.94,2.34] & $4.23\times10^{-6}$\\
MMLU held-out & 11,520 & 314159 & 26.35 & 27.31 & +0.96 & [0.26,1.68] & .00799\\
\bottomrule
\end{tabular}
\end{table}

A smaller validation-only control trains three format-matched response-only verifiers on the same 56/14 MMLU split. On frozen GPQA pools its gains are $+7.07$, $+11.11$, and $+8.08$\,pp; on MMLU confirm-512 they are $+2.34$, $+1.17$, and $+0.98$\,pp, with the latter per-seed intervals crossing zero. This control evaluates small-scale format-matched MCQ supervision separately from the dataset-specific canonical-readout study.

\subsection{Paired constrained decoding}
The four-cell experiment uses Qwen3-1.7B generation and the target-specific canonical verifier at seed 14251. Base trajectories and CD prefixes are paired token-for-token; independently sampled CD trajectories are not substituted. Supported canonical letters are scored once per question and shared across all four cells. The primary contrast family contains \CACC{} minus Base, CD minus Base, and CD+\CACC{} minus CD, with exact McNemar tests and Holm adjustment. Confidence intervals use 10,000 paired question resamples with seed 14251.

\begin{table}[!ht]
\centering\footnotesize
\caption{Fresh-generation four-cell outcomes. $O$, $V|O$, and $F$ are percentages.}
\label{tab:cd-full}
\begin{tabular}{@{}l l r r r@{}}
\toprule Evaluation & Cell & $O$ & $V|O$ & $F$\\\midrule
GPQA ($N=198$) & Base & 20.71 & 82.93 & 17.17\\
& \CACC{} & 51.01 & 45.54 & 23.23\\
& CD & 75.25 & 39.60 & 29.80\\
& CD+\CACC{} & 81.82 & 35.19 & 28.79\\\midrule
MMLU confirm ($N=512$) & Base & 31.45 & 72.05 & 22.66\\
& \CACC{} & 48.83 & 60.00 & 29.30\\
& CD & 62.30 & 48.59 & 30.27\\
& CD+\CACC{} & 68.16 & 41.26 & 28.12\\
\bottomrule\end{tabular}
\end{table}

The Holm-adjusted primary $p$-values, in contrast order, are $.1923$, $.00668$, $.7539$ on GPQA and $.001764$, $.002182$, $.02660$ on MMLU confirm-512. The interactions on $F$ are $-7.07$\,pp (CI $[-14.14,0.00]$) and $-8.79$\,pp (CI $[-12.70,-4.88]$). The additional direct CD-minus-\CACC{} contrasts are $+6.57$\,pp on GPQA (unadjusted $p=.04096$, CI $[+1.01,+12.63]$) and $+0.98$\,pp on MMLU (unadjusted $p=.6718$, CI $[-2.73,+4.50]$). These additional contrasts are outside the primary Holm family.

\subsection{Selector families}
Majority voting and the three LLM scoring variants operate on the original candidate pools. Option-logprob scoring uses an 8B model; candidate judges use 1.7B and 32B models, with the latter Qwen3-32B. The strong judge's candidate-text limit is 2,000 characters and no evaluated candidate is clipped. The MMLU majority row uses the full 12,032 questions, while the model-based MMLU rows use confirm-512; their sample scopes are shown explicitly.

\begin{table}[!ht]
\centering\footnotesize
\caption{Selector-family results on original pools. Accuracies are percentages; $p$ is the exact paired McNemar value for each row, without a table-wide adjustment.}
\label{tab:selector-controls}
\setlength{\tabcolsep}{4pt}
\begin{tabular}{@{}l r l r r r l@{}}
\toprule Evaluation & $N$ & Selector & Base & \CACC{} & $\Delta F$ & $p$\\\midrule
GPQA & 198 & Majority & 16.67 & 19.70 & +3.03 & .2101\\
GPQA & 198 & 8B option-logprob & 17.68 & 28.28 & +10.61 & .00460\\
GPQA & 198 & 1.7B candidate judge & 5.56 & 18.69 & +13.13 & $4.23\times10^{-5}$\\
GPQA & 198 & 32B candidate judge & 6.57 & 23.23 & +16.67 & $1.07\times10^{-6}$\\
MMLU full & 12,032 & Majority & 21.17 & 18.48 & -2.69 & $3.43\times10^{-18}$\\
MMLU confirm & 512 & 8B option-logprob & 32.62 & 32.62 & 0.00 & 1.000\\
MMLU confirm & 512 & 1.7B candidate judge & 18.75 & 17.97 & -0.78 & .7709\\
MMLU confirm & 512 & 32B candidate judge & 23.83 & 25.20 & +1.37 & .5825\\
\bottomrule\end{tabular}
\end{table}

The strong-judge paired intervals are $[+10.10,+23.23]$\,pp on GPQA and $[-2.73,+5.47]$\,pp on MMLU confirm-512. Candidate coverage is identical across selector families within each pool and evaluation scope; final accuracy changes with selection.

\subsection{Five-seed generator study}
Mistral-7B-Instruct-v0.1 generates fresh pools for MMLU confirm-512 at seeds 7, 11, 19, 23, and 31, temperature 0.7, and top-$p$ 0.95. Base generation samples 16 sequences with at most 128 new tokens each. \CACC{} adds four completion sequences with at most 160 new tokens each, and both conditions retain at most eight candidates. The selector is the Qwen3-1.7B MMLU adapter without Mistral-specific retraining. Mean oracle coverage rises from $48.20\%$ to $58.36\%$, and mean final accuracy from $14.22\%$ to $15.94\%$. All five final deltas are positive. Paired-question bootstrap of the fixed-five-seed mean gives $+1.719$\,pp with a 95\% interval $[+0.742,+2.773]$\,pp. This study varies the generator while keeping the selector and retained cap fixed; completion generation remains an additional operation.

\subsection{Format-only intervention and pool diagnostics}
The paired format intervention changes only the final-answer marker, between \texttt{Final answer:} and \texttt{Option <LETTER>}; reasoning text and decoded answer content are fixed. Three format-matched validation-trained verifiers are evaluated on MMLU confirm-512. For the \CACC{} condition, 381 of 512 questions meet the prespecified eligibility rule. Fixed-three-seed effects are averaged within questions before 10,000 question-level bootstrap resamples. The signed margin shift (Final answer minus Option) is $-0.3899$ (95\% CI $[-0.5881,-0.1940]$; exact two-sided sign-test $p=.0027$). Eligible selection flips are $9.89\%$ (CI $[7.96,11.90]\%$). All-question accuracy changes by $-0.065$\,pp (CI $[-0.911,+0.716]$\,pp). The estimand is format sensitivity of selection, not an aggregate accuracy gain.

For the full MMLU majority path, reconstructed snapshot differences identify 1,620 newly covered examples, 4,116 jointly covered examples, and 784 examples losing correct coverage. Of the new hits, 261 convert ($16.11\%$). Conversion on jointly covered examples changes from $57.31\%$ to $47.67\%$. These comparisons describe the two recorded candidate pools rather than a traced generation lineage.

\subsection{Prospective ARC-Challenge validation}
\label{sec:arc-prospective}
We froze the schema wrapper, operative MCQ rules, model, parser, sampling settings, and evaluation protocol before a single paired evaluation on 256 ARC-Challenge test questions \citep{clark2018thinksolvedquestionanswering}. Questions were selected by sorting SHA256(\texttt{CACC-ARC-v1|source\_id}) without labels. Original option labels were mapped in their existing order to A--J. The existing operative payload required zero edits; only the schema/label wrapper was added. Gold labels were excluded from model inputs and accessed only after both predictions were frozen.

Base uses Qwen3-1.7B, 16 samples of at most 128 new tokens, temperature 0.7 and top-$p$ 0.95, retaining at most eight candidates. \CACC{} adds four completions of at most 160 new tokens using the existing benchmark-aware prompt, three context attempts, one protected prefix candidate, and fragment-first replacement. The frozen format-matched MMLU validation-trained verifier (seed 14251, epoch 1) scores candidate-supported canonical letters with one shared score cache. Sampling uses seed 14251, with deterministic per-question offsets to support checkpointed execution. This single-seed study measures reuse on a new benchmark under a fixed selector.

\begin{table}[!ht]
\centering\small
\caption{Prospective ARC-Challenge results ($N=256$). Oracle ($O$), conditional conversion ($V$), and final accuracy ($F$) are percentages.}
\label{tab:arc-prospective}
\begin{tabular}{lrrr}\toprule Condition & $O$ & $V$ & $F$\\\midrule
Base & 85.16 & 82.11 & 69.92 \\
\CACC{} & 91.02 & 77.68 & 70.70 \\
\bottomrule\end{tabular}
\end{table}

Final accuracy changes by $+0.78$\,pp (10,000 paired-question bootstrap resamples; 95\% CI $[-2.34,+3.91]$\,pp; exact two-sided McNemar $p=0.8036$). The final-accuracy difference is not statistically significant.

The cumulative single-GPU session ledger records 0.548 GPU allocation wall-clock hours, including runtime preparation probes and a conservative charge for a timed-out probe; this is not a measurement of GPU busy time or energy. The recorded implementation-and-runtime setup window spans 19.1 minutes from the first saved preflight script to protocol freeze, including compatibility checks and orchestration. This is an agent workflow wall-clock interval; human adaptation time was not measured.

\subsection{Release provenance and adaptation measurement}
The public implementation is maintained at \url{https://github.com/drongzzz0/CACC-nips}. The inspected release version is commit \texttt{8fdf207}. Its per-row ledger distinguishes original artifacts, reproduced display ratios, reconstructed reruns, and missing prediction-level provenance. Some stronger-proposer rows have reconstructed reruns rather than the exact original candidate streams; these are documented as separate protocols. The original MMLU Base pair is unavailable for a new paired test of the historical $-0.85$\,pp contrast. New controls above use their own paired records and do not substitute for that pair.

A source audit finds a common operative MCQ payload across GPQA and MMLU-Pro, with benchmark-specific wrappers. This is retrospective implementation reuse. Appendix~\ref{sec:arc-prospective} adds a prospective study with the benchmark, rulebook, model, parser, and metric frozen before evaluation; agent setup time and compute are recorded separately from unmeasured human effort.

\paragraph{Broader impact.}
Improving smaller models can support local and resource-constrained reasoning applications. Better answer formatting also makes an incorrect answer easier to consume, so parser validity should be monitored together with correctness. The framework does not provide correctness guarantees or replace application-specific validation. Release materials should respect benchmark licenses and access restrictions, including the handling of GPQA content.

\end{document}